\pdfoutput=1

\documentclass[11pt]{article}

\usepackage[final]{acl}

\usepackage{times}
\usepackage{latexsym}

\usepackage[T1]{fontenc}

\usepackage{pgfplotstable}
\usepackage{longtable}
\usepackage{booktabs}

\usepackage{amsmath,amsfonts,amssymb}
\usepackage{amsthm}
\usepackage{graphicx}
\usepackage{subcaption}
\usepackage{bm}
\usepackage{cases}
\usepackage{multirow}
\usepackage{booktabs}
\usepackage{mathtools}
\usepackage{color}
\usepackage{stmaryrd}
\usepackage{pifont}
\usepackage{changes}

\usepackage{enumitem}

\usepackage[utf8]{inputenc}

\usepackage{microtype}

\usepackage{inconsolata}

\usepackage{graphicx}
\usepackage{needspace}

\title{Task-Stratified Knowledge Scaling Laws for Post-Training Quantized Large Language Models}

\author{
  Chenxi Zhou\textsuperscript{1,2},
  Pengfei Cao\textsuperscript{2,3}\thanks{Corresponding authors},
  Jiang Li\textsuperscript{4},
  Bohan Yu\textsuperscript{1,2},
  Jinyu Ye\textsuperscript{2},
  Jun Zhao\textsuperscript{2,3},
  Kang Liu\textsuperscript{2,3}\footnotemark[1]
  \\
  \\
  \textsuperscript{1}School of Advanced Interdisciplinary Sciences, University of Chinese Academy of Sciences \\
  \textsuperscript{2}The Key Laboratory of Cognition and Decision Intelligence for Complex Systems, \\ Institute of Automation, Chinese Academy of Sciences \\
  \textsuperscript{3}School of Artificial Intelligence, University of Chinese Academy of Sciences \\
  \textsuperscript{4}College of Computer Science, Inner Mongolia University \\
  \texttt{zhouchenxi2025@ia.ac.cn, \{pengfei.cao, jzhao, kliu\}@nlpr.ia.ac.cn} \\[1.5em]
}

\begin{document}
\maketitle

\begin{abstract}
Post-Training Quantization (PTQ) is a critical strategy for efficient Large Language Models (LLMs) deployment. However, existing scaling laws primarily focus on general performance, overlooking crucial fine-grained factors and how quantization differentially impacts diverse knowledge capabilities. To address this, we establish Task-Stratified Knowledge Scaling Laws. By stratifying capabilities into memorization, application, and reasoning, we develop a framework that unifies model size, bit-width, and fine-grained factors: group size and calibration set size. Validated on 293 diverse PTQ configurations, our framework demonstrates strong fit and cross-architecture consistency. It reveals distinct sensitivities across knowledge capabilities: reasoning is precision-critical, application is scale-responsive, and memorization is calibration-sensitive. We highlight that in low-bit scenarios, optimizing these fine-grained factors is essential for preventing performance collapse. These findings provide an empirically-backed foundation for designing knowledge-aware quantization strategies.

\end{abstract}


\section{Introduction}
\label{sec:1_intro}

Large language models (LLMs) have achieved impressive performance across diverse tasks~\cite{guo_evaluating_2023}, but their growing scale poses deployment challenges due to high memory and computational costs~\cite{zhu_survey_2024, lang_comprehensive_2024}. Post-training quantization (PTQ) emerges as a practical solution by compressing LLMs without expensive retraining~\cite{yao_comprehensive_2023}. A recent study shows that nearly 70\% of quantization-related research since 2022 has focused on PTQ for LLMs~\cite{zhao_new_2025}.

Despite the widespread use of PTQ, a comprehensive understanding of how LLM performance is precisely impacted under quantization remains elusive. Current evaluations offer general insights, such as performance cliffs below 4-bit precision~\cite{li_icml_2024} and task-specific sensitivities~\cite{marchisio_emnlp_2024, liu_new_2025}. However, these studies typically lack a systematic and predictive framework. This deficiency makes it difficult for practitioners to make informed decisions when configuring PTQ strategies. To this end, some researchers have initiated the exploration of scaling laws for quantized models, aiming to establish relationships between model performance and factors, such as model size or bit-width~\cite{ouyang_low-bit_2024, kumar_iclr_2025, xu_scaling_2024}. Such scaling laws enable the prediction of post-quantization performance. However, they still have two notable limitations:

\textbf{1) The role of fine-grained PTQ factors is overlooked.} 
Current studies predominantly focus on factors like model size, bit-width, and pre-training data volume~\cite{ouyang_low-bit_2024, kumar_iclr_2025}. In contrast, tunable parameters inherent in widely adopted algorithms (e.g., GPTQ~\cite{frantar_iclr_2023}), such as group size~\cite{elangovan_bcq_2025} and calibration set size~\cite{zhang_selectq_2025}, are often treated as constants. However, our empirical observations reveal that these fine-grained parameters are decisive factors for maintaining model capabilities, especially under low-bit quantization.

\textbf{2) The impact of quantization on diverse knowledge capabilities remains underexplored.} 
Existing scaling laws mainly focus on the overall performance of quantized LLMs, often overlooking the fact that LLMs possess diverse knowledge capabilities. This is critical as they rely on core capabilities, ranging from memorization to application and reasoning, to support diverse downstream tasks~\cite{wang_emnlp_2024, yu_iclr_2024}. Crucially, these capabilities are hypothesized to exhibit divergent sensitivities to quantization due to their distinct underlying mechanisms, which general scaling laws fail to capture.

To address these limitations, we conduct an extensive empirical investigation to establish \textbf{Task-Stratified Knowledge Scaling Laws} for post-training quantized LLMs. Specifically, this involves: 
1) \emph{systematically incorporating model size, bit-width, calibration set size, and group size into a unified power-law framework}; and 2) \emph{comprehensively investigating the impact of quantization configurations on the diverse knowledge capabilities of LLMs}. 
Validated on 293 diverse PTQ configurations spanning the Qwen3 and Llama-3 families, our framework demonstrates a strong fit and cross-architecture universality. We reveal that different knowledge capabilities exhibit distinct sensitivities to quantization variables. Specifically, while reasoning is bottlenecked by precision (bit-width and group size), knowledge application scales significantly with model size, and memorization is particularly sensitive to calibration set size. 
Furthermore, we highlight that under low-bit quantization, smaller group sizes and sufficient calibration data are no longer optional but essential to prevent performance collapse.



In summary, our contributions are twofold:
\begin{itemize} [topsep=1pt, itemsep=3pt, parsep=1pt]
    \item We establish the first task-stratified knowledge scaling laws for PTQ. Our unified framework incorporates model size and bit-width alongside crucial fine-grained factors (group size and calibration set size), and models diverse knowledge capabilities separately.
    \item We empirically reveal divergent sensitivities across knowledge capabilities (memorization, application, and reasoning) to quantization, and highlight that optimizing fine-grained factors is essential for preventing performance collapse under low-bit scenarios.
\end{itemize}


\section{Related Work}
\label{sec:2_related_work}

\subsection{Post-Training Quantization of LLMs}
Post-Training Quantization (PTQ) has emerged as a dominant strategy for LLM compression, offering superior efficiency over Quantization-Aware Training (QAT) by eliminating retraining~\cite{lang_comprehensive_2024, hasan_optimizing_2024}. While PTQ methods vary widely, they generally balance compression and performance via sophisticated calibration techniques~\cite{williams_acl_2024, ji_beware_2024}. 

Among these, optimization-based approaches like GPTQ~\cite{frantar_iclr_2023} have become industry standards. GPTQ leverages second-order information (Hessian matrix) and calibration data to minimize quantization error layer-by-layer. Crucially, the performance of such methods is intricately tied to hyperparameters like calibration set size and group granularity~\cite{zhang_selectq_2025, elangovan_bcq_2025}. However, prior works typically treat these as static settings rather than dynamic scaling variables, leaving their systematic impact on model capabilities underexplored.


\subsection{Scaling Laws for Quantized LLMs}

Neural scaling laws provide a predictive framework linking model performance to resources. Pioneering works by \citet{kaplan_scaling_2020} and \citet{hoffmann_training_2022} establish that uncompressed LLM performance follows power laws with model size, training tokens, and training compute.

Recently, this framework has been extended to the quantization domain. For instance, \citet{ouyang_low-bit_2024} investigate scaling laws for quantization-induced degradation (QiD), linking QiD to training data volume, model size, and bit-width. \citet{kumar_iclr_2025} explore the interplay between training precision and PTQ precision. \citet{sun_icml_2025} explore the scaling behavior of floating-point representation structures during the training phase. Furthermore, \citet{xu_scaling_2024} attempt to build predictive models for post-PTQ quality considering various factors.  

Despite these advancements, prior works primarily focus on generic performance metrics, overlooking how varying quantization configurations differentially impact distinct knowledge capabilities. The lack of a unified framework incorporating fine-grained factors leaves the scaling dynamics of diverse capabilities largely unquantified.

\section{Task-Stratified Knowledge Scaling Laws for PTQ LLMs}

\subsection{Task Capability Definitions for Quantization Analysis}
\label{sec:3.1}

\begin{figure}[t!]
    \centering
    \includegraphics[width=1.0\linewidth]{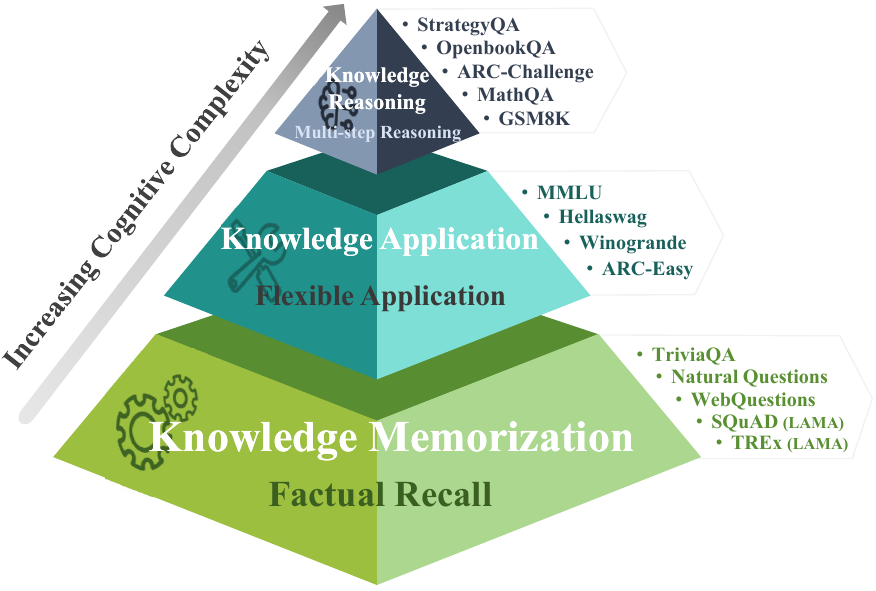}
    \caption{Overview of the task-stratified knowledge taxonomy defined in this study.}
    \label{fig:pyramid}
\end{figure}

To systematically investigate the impact of PTQ on LLMs, we refine the knowledge capability taxonomy into three hierarchical levels of increasing cognitive complexity, as illustrated in Figure~\ref{fig:pyramid}: \textit{knowledge memorization}, \textit{knowledge application}, and \textit{knowledge reasoning}.

This stratification draws from Bloom's Taxonomy~\cite{krathwohl_revision_2002, huber_coling_2025}, its adaptation for LLM benchmarks (e.g., KoLA~\cite{yu_iclr_2024}), and recent studies on knowledge mechanisms in LLMs~\cite{wang_emnlp_2024}. We posit that these knowledge capabilities exhibit divergent sensitivities to quantization, necessitating a task-stratified scaling analysis.

\noindent\textbf{Level 1: Knowledge Memorization (KM).}
Aligning with Bloom's \textit{Remembering} level, this capability refers to an LLM's ability to accurately store and recall specific factual knowledge learned during pre-training. Tasks at this level are characterized by an ``exact lookup'' nature, where the model must recall precise facts (e.g., names, dates) from the internal knowledge base without complex contextual transformation.

\noindent\textbf{Level 2: Knowledge Application (KA).}
Combining Bloom's \textit{Understanding} and \textit{Applying} levels, KA transcends static storage, focusing on comprehending inquiries and leveraging internalized knowledge to formulate appropriate answers. Unlike simple recall, this level requires the model to understand the context and apply generalized knowledge to specific scenarios, emphasizing flexible application rather than strict factual knowledge lookup.

\noindent\textbf{Level 3: Knowledge Reasoning (KR).}
Aligning with Bloom's deep thinking skills (primarily \textit{Analyzing}~\cite{huber_coling_2025}), KR involves complex cognitive processes including multi-step logic, mathematical problem-solving, and chain-of-thought deduction~\cite{wei_chain--thought_2022}. Unlike application, complex reasoning requires the model to construct multi-step logical chains to handle novel scenarios beyond simple pattern matching. 

Based on this stratification, we aim to construct distinct scaling laws for each level, predicting how PTQ configurations impact diverse knowledge capabilities.

\subsection{Factors under Investigation}
\label{sec:3.2_factors}

To establish task-stratified scaling laws, we focus on four key factors governing the quantization process.
Fundamentally, PTQ compresses a model of size $N$ by mapping high-precision weights $\mathbf{W}$ to $B$-bit representations $\hat{\mathbf{W}}$.
This process typically aims to minimize the reconstruction error $\|\mathbf{W}\mathbf{X} - \hat{\mathbf{W}}\mathbf{X}\|_F^2$ on calibration inputs $\mathbf{X}$ (with set size $C_b$).
Furthermore, the quantization granularity is determined by the group size $G$, which defines the block size of weights sharing the same quantization scale (and zero-point).
We examine the scaling behaviors of these factors below:


\begin{figure*}[t]
  \centering
  \includegraphics[width=0.95\textwidth]{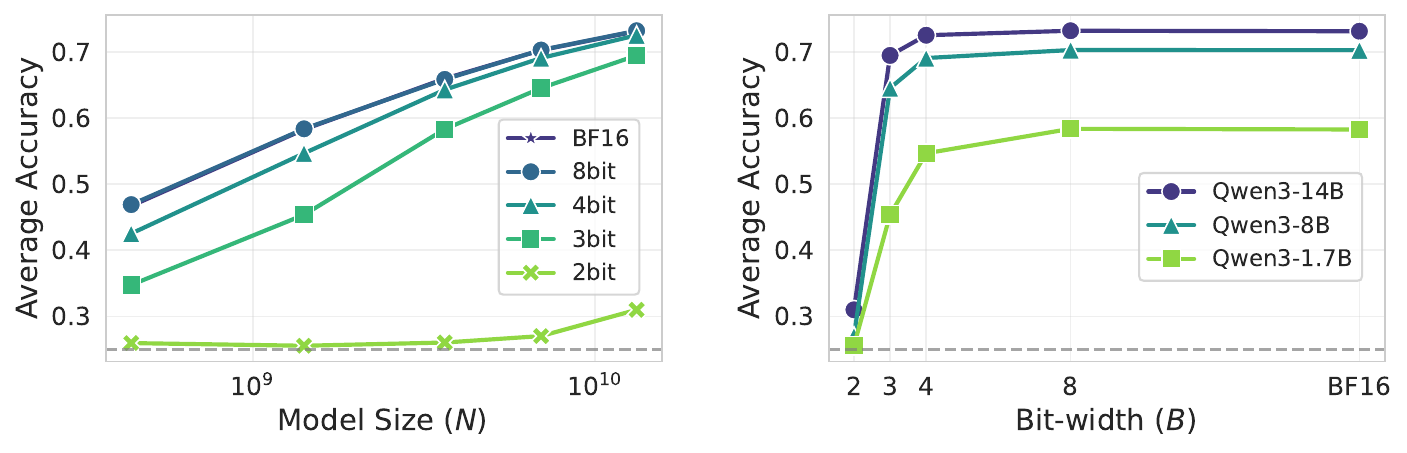} 
  \caption{Scaling trends of Model Size ($N$) and Bit-width ($B$) for Qwen3 models ($C_b=128, G=128$). Accuracy is averaged across five representative 4-choice tasks: Hellaswag, ARC-e/c, MMLU, and OpenbookQA. The dashed grey line represents the random baseline (0.25). (BF16 and 8-bit curves visually overlap).}
  \label{fig:macro_trends}
\end{figure*}

\textbf{(1) Model Size ($N$):} 
Defined as the total number of non-embedding parameters~\cite{ouyang_low-bit_2024}, model size determines representational capacity and robustness to quantization noise. Figure~\ref{fig:macro_trends} (left) confirms that accuracy consistently increases with model size across most bit-widths, following a power-law trend as in full-precision models~\cite{kaplan_scaling_2020, hoffmann_training_2022}. However, the 2-bit models remain near the random baseline and improve only slightly at large scales, deviating markedly from higher-precision trends.

\textbf{(2) Bit-width ($B$):} 
As shown in Figure~\ref{fig:macro_trends} (right), we observe a sharp recovery: performance rises steeply from the random baseline at 2-bit to a usable level at 3-bit, before saturating near BF16 performance at higher bit-widths. This observation highlights the non-linear impact of bit-width on model capabilities.


\begin{figure*}[t]
  \centering
  \includegraphics[width=0.95\textwidth]{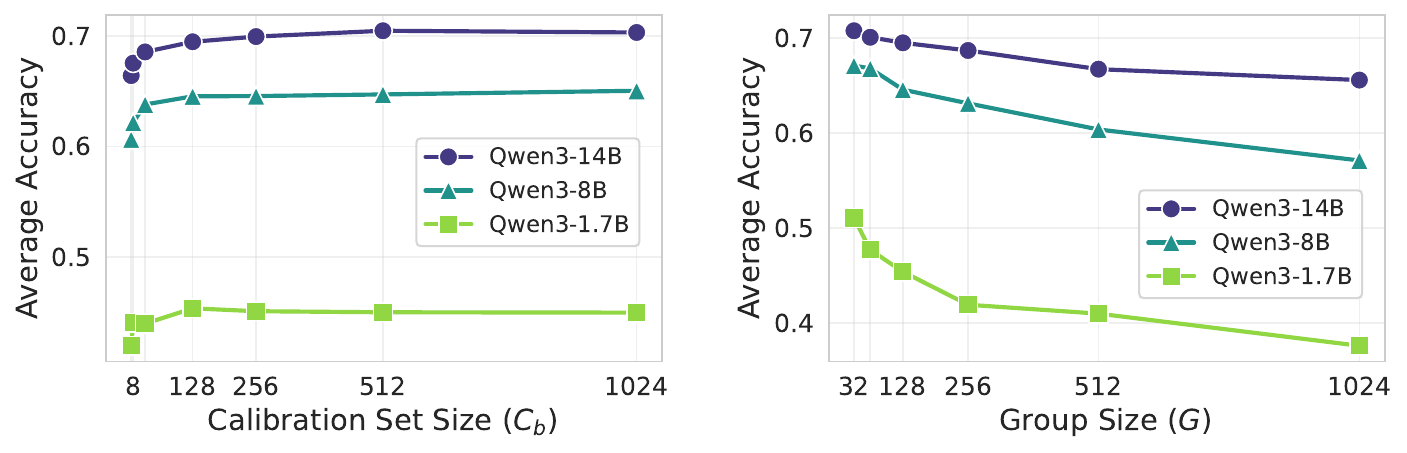}
  \caption{Scaling trends of Calibration Set Size ($C_b$) and Group Size ($G$) under 3-bit quantization. Benchmarks are the same as in Figure~\ref{fig:macro_trends}. 
  (Left) Impact of $C_b$ with fixed $G=128$.
  (Right) Impact of $G$ with fixed $C_b=128$.}
  \label{fig:micro_trends}
\end{figure*}

\textbf{(3) Calibration Set Size ($C_b$):}
While the importance of calibration data is acknowledged~\cite{zhang_selectq_2025, williams_acl_2024, ji_beware_2024}, its systematic scaling behavior remains under-explored. As shown in Figure~\ref{fig:micro_trends} (left), increasing $C_b$ improves accuracy, but the benefits saturate at larger sizes. This non-linear saturation motivates its inclusion as a key factor to quantify its impact on knowledge preservation.

\textbf{(4) Group Size ($G$):} 
Group size serves as a trade-off between compression ratio and error compensation. Figure~\ref{fig:micro_trends} (right) demonstrates a pronounced inverse relationship: smaller group sizes (e.g., 32, 64) mitigate accuracy loss via finer-grained quantization, whereas larger groups (e.g., 1024) cause obvious degradation. This confirms that $G$ acts as a critical granularity regulator in PTQ.


\subsection{Scaling Law Formulation and Fitting Method}


\subsubsection{Task-Stratified Scaling Law}

To quantitatively model the impact of quantization configurations on knowledge capabilities, we propose a unified multiplicative power-law function. The performance metric, denoted as the negative log-adjusted accuracy, is modeled as follows:
\begin{equation} \label{eq:scaling-law}
\begin{split}
    -\ln(\mathrm{Acc}_{\text{adj}}) &= A_{\text{task}} \cdot N^{\alpha_{\text{task}}} (\log_2 B)^{\beta_{\text{task}}} \\
    &\quad (\log_2 C_b)^{\gamma_{\text{task}}} G^{\delta_{\text{task}}},
\end{split}
\raisetag{20pt}  
\end{equation}
where $A_{\text{task}}$ is a task-specific constant scaling coefficient. The exponents $\alpha_{\text{task}}, \beta_{\text{task}}, \gamma_{\text{task}}, \text{and } \delta_{\text{task}}$ are task-specific scaling parameters, quantifying the sensitivity of performance on that task type to each respective factor. 

Note that since higher performance corresponds to a lower value of $-\ln(\mathrm{Acc}_{\text{adj}})$, we expect negative exponents for resource-related factors ($N, B, C_b$), as scaling them up reduces this ``loss'' metric. Conversely, we anticipate a positive exponent for group size ($G$), since a larger group size implies coarser quantization granularity, which typically degrades performance (increases the ``loss'').

\paragraph{Theoretical Support.} The adoption of this functional form is based on two key foundations. 
First, the multiplicative power-law structure successfully describes how neural networks scale, capturing the relationship between influential factors and model performance~\cite{kaplan_scaling_2020, hoffmann_training_2022}. 
Second, we fit the negative natural logarithm of the adjusted accuracy instead of raw accuracy. As highlighted by \citet{schaeffer_why_2025}, downstream metrics like accuracy are bounded in $[0,1]$ and exhibit complex non-linear behaviors that are difficult to fit directly. Transforming accuracy into an unbounded ``loss-like'' space ($-\ln(\mathrm{Acc})$) restores the monotonic, convex properties required for robust modeling~\cite{krajewski_revisiting_2025}. This form also allows the exponents to be understood as elasticities, quantifying the sensitivity of performance to relative changes in each factor.

\paragraph{Adjustment for Diverse Task Baselines.} 
Our evaluation spans a diverse three-layer knowledge taxonomy where random guessing baselines ($\mathrm{Acc}_{\text{random}}$) vary significantly. For instance, generative tasks in knowledge memorization have a baseline approaching zero, whereas multiple-choice tasks in knowledge application have a baseline of 0.25 or 0.5. To eliminate this bias and ensure a unified scaling metric across different task types, we use the baseline-adjusted accuracy instead of raw accuracy:
\begin{equation}
    \mathrm{Acc}_{\text{adj}} = \frac{\mathrm{Acc} - \mathrm{Acc}_{\text{random}}}{1 - \mathrm{Acc}_{\text{random}}} \,.
\end{equation}
This adjustment ensures that $\mathrm{Acc}_{\text{adj}}$ reflects knowledge gain over random guessing, enabling consistent comparison in our task-stratified analysis.

\subsubsection{Illustration for Logarithmic Transformation of $C_b$ and $B$}

As introduced in Eq.~\ref{eq:scaling-law}, we apply a logarithmic transformation ($\log_2$) to both calibration set size ($C_b$) and bit-width ($B$) to explicitly model their non-linear ``diminishing returns'' on model accuracy. Specifically, as observed in our preliminary experiments (Figure~\ref{fig:macro_trends} and~\ref{fig:micro_trends}), initial increases in $C_b$ or $B$ yield substantial performance gains, but these benefits progressively diminish as the values become larger. The logarithmic transformation linearizes this saturation behavior, ensuring robust fitting across the effective range. This modeling choice aligns with prior work suggesting that the utility of additional calibration data~\cite{williams_acl_2024} and increased bit-width~\cite{li_icml_2024} often follows such a non-linear pattern.

\subsubsection{Fitting Method}

To robustly estimate the coefficients ($A_{\text{task}}, \alpha_{\text{task}}, \beta_{\text{task}}, \gamma_{\text{task}}, \delta_{\text{task}}$), we transform the multiplicative scaling law into a linear form by taking the natural logarithm of both sides of Eq.~\ref{eq:scaling-law}:
\begin{equation} \label{eq:fitting_linear}
\begin{split}
    \ln(-\ln(\mathrm{Acc}_{\text{adj}})) & = \ln A_{\text{task}} \!+\! \alpha_{\text{task}} \ln N \\ 
    & +\! \beta_{\text{task}} \ln(\log_2 B) \\
    & +\! \gamma_{\text{task}} \ln(\log_2 C_b) \!+\! \delta_{\text{task}} \ln G .
\end{split}
\raisetag{2.5\baselineskip} 
\end{equation}
We employ Ordinary Least Squares (OLS) linear regression~\cite{zdaniuk_ordinary_2014} on this log-log data, filtering out collapsed configurations ($\mathrm{Acc}_{\text{adj}} \leq 0.01$) to ensure numerical stability (Appendix~\ref{subsec:numerical_stab}). Compared to direct Non-linear Least Squares (NLS) optimization, this linearized approach offers a closed-form solution and ensures convexity, avoiding local optima~\cite{sengupta_compression_2025}.



To rigorously evaluate the model's explanatory power, we employ the Adjusted $R^2$ statistic (Appendix~\ref{appendix:r2}). We report this metric in two spaces: (1) the log-space ($\ln(-\ln(\mathrm{Acc}_{\text{adj}}))$) to assess regression quality, and (2) the original space ($\mathrm{Acc}_{\text{adj}}$) to validate practical predictive capability. Furthermore, we utilize Mean Absolute Error (MAE) to verify absolute accuracy and extrapolation robustness (Appendix~\ref{subsec:mae_results}).


\section{Experiments}

\subsection{Experimental Setup}
\label{sec:exp_setup}

We design a comprehensive setup to evaluate how PTQ parameters affect distinct knowledge capabilities. The implementation details, along with the rationale for benchmark stratification, are provided in Appendix~\ref{sec:appendix_exp}.

\paragraph{Models.} 
We primarily study the Qwen3 family~\cite{yang_qwen3_2025}, chosen for its recency and the broad coverage of available model sizes, which facilitates robust scaling analysis. We use five sizes for scaling law fitting: 0.6B, 1.7B, 4B, 8B, and 14B. Additionally, Qwen3-32B is reserved to validate the extrapolation of our proposed laws.

\paragraph{Benchmarks.} 

We evaluate diverse knowledge capabilities using 14 representative benchmarks aligned with the taxonomy defined in Section~\ref{sec:3.1}.
\begin{itemize}[leftmargin=*, topsep=1pt, itemsep=1pt, parsep=1pt]
    \item \textbf{L1 (KM).} Assessed via benchmarks requiring exact facts recall, including TriviaQA~\cite{joshi_triviaqa_2017}, Natural Questions~\cite{kwiatkowski_natural_2019}, WebQuestions~\cite{berant_semantic_2013}, and the TREx and SQuAD subsets from LAMA~\cite{petroni_emnlp_2019}.
    \item \textbf{L2 (KA).} Evaluated on tasks focusing on flexible knowledge application, specifically Hellaswag~\cite{zellers_hellaswag_2019}, Winogrande~\cite{sakaguchi_winogrande_2021}, MMLU~\cite{hendrycks_measuring_2021}, and ARC-Easy~\cite{clark_think_2018}.
    \item \textbf{L3 (KR).} Tested using multi-step reasoning datasets, namely StrategyQA~\cite{geva_did_2021}, OpenbookQA~\cite{mihaylov_can_2018}, ARC-Challenge~\cite{clark_think_2018}, GSM8K~\cite{cobbe_training_2021}, and MathQA~\cite{amini_mathqa_2019}.
\end{itemize}


\paragraph{Quantization Strategy.}
Establishing robust scaling laws requires systematic sweeps over multiple quantization variables. We employ GPTQ~\cite{frantar_iclr_2023} because it is the most widely adopted weight-only PTQ method, and its mature libraries readily support the flexible configurations essential for our analysis. In contrast, implementations of alternative methods (e.g., AWQ~\cite{lin_mlsys_2024}, QuIP~\cite{chee_nips_2023}) often restrict accessible bit-widths or architectures. We apply a targeted sampling strategy to different compression zones. In the effective compression zone (3/4-bit), we execute a full grid search ($C_b \in \{8, 32, 128, 1024\}$, $G \in \{32, 64, 128, 1024\}$) to capture fine-grained sensitivities. Conversely, 8-bit configurations are fixed ($C_b=128, G=128$) due to marginal variance, and 2-bit is excluded from overall fitting to strictly preserve power-law assumptions.


\subsection{Validation of the Unified Scaling Law}
\label{sec:general_scaling_law}

We first validate our unified scaling law on aggregated performance across all knowledge levels, offering an overall view of how PTQ factors influence general model performance.

\subsubsection{Goodness-of-Fit and Ablation Analysis}
\label{sec:fit_analysis}

We perform an ablation study to quantify the contribution of each factor. The results, summarized in Table~\ref{tab:ablation_fitting} and visualized in Figure~\ref{fig:overall_fit}, reveal several key insights regarding factor importance:

\begin{figure}[t]
    \centering
    \includegraphics[width=\columnwidth]{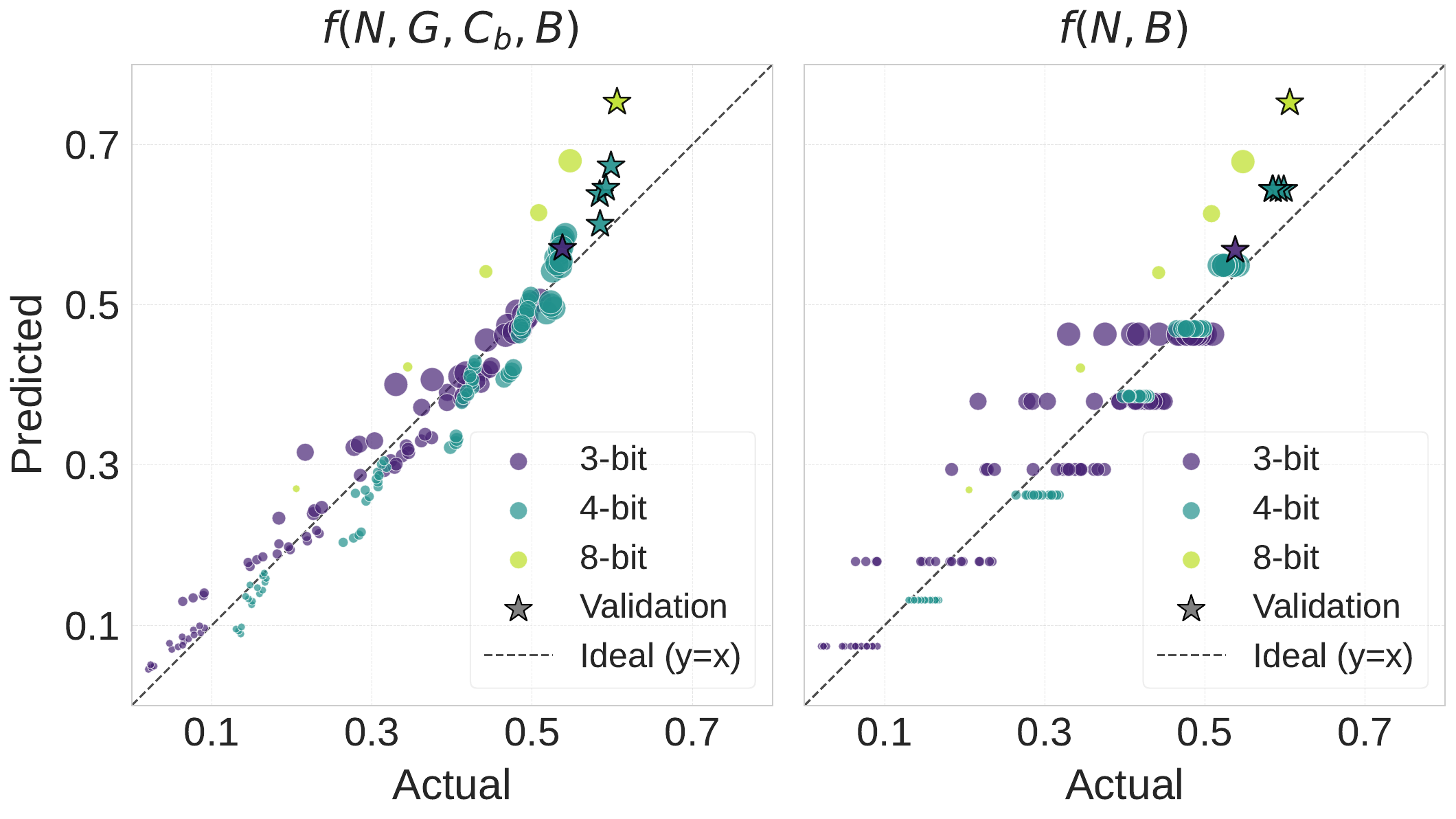}
    \caption{Goodness-of-fit: Predicted vs. actual adjusted accuracy for (Left) our proposed four-factor law ($N, B, C_b, G$) and (Right) the baseline ($N, B$). Points are colored by bit-width ($B$) and sized by model size ($N$). Stars ($\star$) denote the validation data (Qwen3-32B). Dashed line represents ideal prediction.}
    \label{fig:overall_fit}
\end{figure}


\begin{table*}[t]
  \centering
  \small
  \renewcommand{\arraystretch}{1.5}
  \setlength{\tabcolsep}{8pt}
  \begin{tabular*}{\textwidth}{@{\extracolsep{\fill}} l l c c @{}}
    \toprule
    \textbf{Formulation} & \textbf{Fitted Function} & \textbf{Adj. $R^2_{\mathcal{L}}$} & \textbf{Adj. $R^2_{\mathcal{O}}$} \\
    \midrule
    $f(N, B, C_b, G)$ & 
    $3.98 \times 10^3 \, N^{-0.359} \, (\log_2 B)^{-1.067} \, (\log_2 C_b)^{-0.032} \, G^{0.073}$ & 
    \textbf{0.9425} & \textbf{0.9475} \\
    $f(N, B)$ & 
    $5.39 \times 10^3 \, N^{-0.359} \, (\log_2 B)^{-1.071}$ & 
    0.9038 & 0.9125 \\
    $f(N, B, G)$ & 
    $3.77 \times 10^3 \, N^{-0.359} \, (\log_2 B)^{-1.071} \, G^{0.073}$ & 
    0.9420 & 0.9466 \\
    $f(N, B, C_b)$ & 
    $5.69 \times 10^3 \, N^{-0.359} \, (\log_2 B)^{-1.067} \, (\log_2 C_b)^{-0.032}$ & 
    0.9041 & 0.9131 \\
    \bottomrule
  \end{tabular*}
  \caption{Ablation analysis of the scaling law formulation modeling $-\ln(\mathrm{Acc}_{\text{adj}})$. Adj. $R^2_{\mathcal{L}}$ and Adj. $R^2_{\mathcal{O}}$ denote the adjusted $R^2$ in the log-transformed and original accuracy spaces, respectively. The full formulation achieves the highest explanatory power, accurately capturing the variance across 165 fitted configurations.}
  \label{tab:ablation_fitting}
\end{table*}

\noindent\textbf{(1) The comprehensive model achieves superior fit.} 
The full four-factor model yields the highest Adj.$R^2_{\mathcal{O}}$ of 0.9475, indicating robust predictive capability. As shown in Figure~\ref{fig:overall_fit} (Left), empirical data points tightly cluster around the ideal diagonal, while the held-out large-scale models (stars) validate extrapolation potential.


\noindent\textbf{(2) Foundational role of $N$ and $B$.} 
The baseline model considering only model size ($N$) and bit-width ($B$) achieves a respectable foundation (Adj.$R^2_{\mathcal{O}} = 0.9125$). The large negative exponents for $N$ ($-0.359$) and $\log_2 B$ ($-1.067$) confirm them as primary drivers for reducing the ``loss'' metric ($-\ln(\mathrm{Acc})$). However, the visible scatter in Figure~\ref{fig:overall_fit} (Right) and the explanatory gap compared to the full formulation ($0.91$ vs. $0.95$) indicate that neglecting granular parameters fails to capture critical performance variations.

\noindent\textbf{(3) Significance of fine-grained factors ($G$ and $C_b$).} 
Combining group size ($G$) and calibration set size ($C_b$) bridges the performance gap. Notably, adding $G$ alone boosts the $Adj.R^2_{\mathcal{O}}$ significantly to 0.9466, identifying it as a critical regulator. While adding $C_b$ yields a marginal statistical gain overall (consistent with saturation effects), it remains indispensable for stability in low-bit scenarios, as discussed below.

\begin{table*}[!t]
  \centering
  \small
  \renewcommand{\arraystretch}{1.3}
  \setlength{\tabcolsep}{10pt}
  \begin{tabular}{l c c c c c c c}
    \toprule
    \multirow{2}{*}{\textbf{Task Level}} & \multirow{2}{*}{\textbf{Const ($A$)}} & \multicolumn{4}{c}{\textbf{Scaling Exponents (Sensitivity)}} & \multicolumn{2}{c}{\textbf{Goodness-of-Fit}} \\
    \cmidrule(lr){3-6} \cmidrule(lr){7-8}
     & & \textbf{$\alpha(N)$} & \textbf{$\beta(B)$} & \textbf{$\gamma(C_b)$} & \textbf{$\delta(G)$} & \textbf{Adj. $R^2_{\mathcal{L}}$} & \textbf{Adj. $R^2_{\mathcal{O}}$} \\
    \midrule
    \textbf{General} & $3.98 \times 10^3$ & -0.359 & -1.067 & -0.032 & 0.073 & 0.9425 & 0.9475 \\
    \midrule
    \textbf{L1: Memorization (KM)} & $2.08 \times 10^3$ & -0.315 & -0.964 & \textbf{-0.040} & 0.064 & 0.9341 & 0.9350 \\
    \textbf{L2: Application (KA)}  & $7.37 \times 10^3$ & \textbf{-0.409} & -0.982 & -0.023 & 0.069 & 0.9550 & 0.9626 \\
    \textbf{L3: Reasoning (KR)}    & $1.27 \times 10^4$ & -0.405 & \textbf{-1.356} & -0.034 & \textbf{0.087} & 0.9156 & 0.9218 \\
    \bottomrule
  \end{tabular}
  \caption{Fitted scaling parameters for task-stratified scaling laws. The model form is $-\ln(\mathrm{Acc}_{\text{adj}}) = A \cdot N^{\alpha} (\log_2 B)^{\beta} (\log_2 C_b)^{\gamma} G^{\delta}$.}
  \label{tab:stratified_params}
\end{table*}

\subsubsection{Parameter Sensitivity in Low-Bit Scenarios}
\label{sec:sensitivity}

\begin{figure}[t]
    \centering
    \includegraphics[width=0.9\columnwidth]{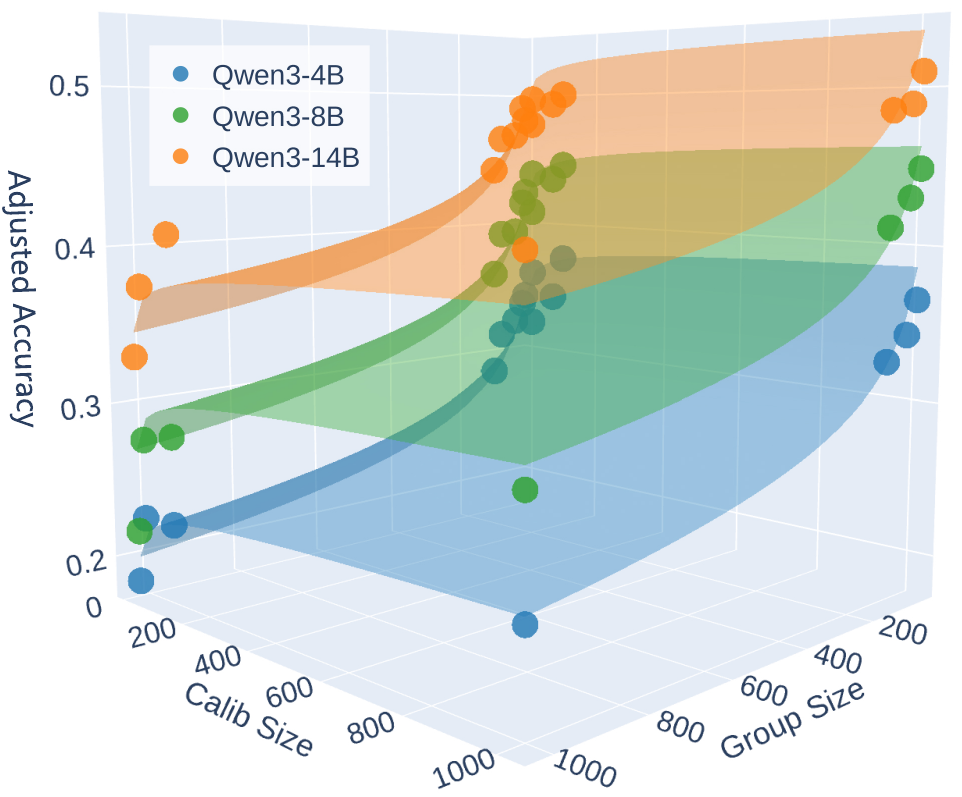}
    \caption{
      Performance surface of the General Scaling Law in the 3-bit region ($\mathrm{Acc}_{\text{adj}} = \exp\left[ - 966.56 \cdot N^{-0.322} (\log_2 C_b)^{-0.103} G^{0.117} \right]$, Adj.$R^2_{\mathcal{O}} = 0.97$). Points represent empirical data.}
    \label{fig:3d_sensitivity}
\end{figure}

While the general model captures global trends, it obscures the nuanced behaviors in the critical 3-bit region. As illustrated in Figure~\ref{fig:3d_sensitivity}, the ``Effective Compression Zone'' exhibits a dramatic sensitivity amplification to fine-grained parameters. 

Specifically, when fitting solely to 3-bit data, the elasticity of calibration data ($C_b$) triples ($|-0.032| \to |-0.103|$), confirming its shift from a diminishing factor to a critical constraint. Simultaneously, the group size ($G$) coefficient surges ($0.073 \to 0.117$), indicating that coarse grouping becomes penalizing at lower precisions. These trends further intensify in the 2-bit region, as we will discuss in Section~\ref{sec:2bit_phase_transition}.

\subsection{Task-Stratified Scaling Laws}
\label{sec:stratified}

While the general scaling law provides a macroscopic view, it inevitably masks the distinct scaling behaviors of different knowledge capabilities. To dissect these nuances, we derive separate scaling laws for the three knowledge levels: knowledge memorization, application, and reasoning. We fit the full four-variable formulation to each task level independently. Detailed ablation studies for each level are provided in Appendix~\ref{subsec:ablation_qwen3}.

\subsubsection{Heterogeneous Sensitivity Analysis}
\label{sec:coefficients}

Table~\ref{tab:stratified_params} details the fitted parameters for each knowledge level (standard errors and 95\% confidence intervals are provided in Appendix~\ref{subsec:stats_qwen3} to confirm statistical significance). As shown, all stratified formulations achieve high goodness-of-fit, confirming the universality of the proposed power-law formulation. However, a cross-comparison of the exponents reveals divergent sensitivities to quantization.


\noindent\textbf{(1) Reasoning (KR) is Precision-Critical.} 
L3 tasks exhibit the highest sensitivity to bit-width ($\beta = -1.356$) and group granularity ($\delta = 0.087$). Notably, the bit-width sensitivity exceeds that of KM and KA by nearly 40\%. This supports the hypothesis that reasoning relies on long-chain logical deductions, where quantization noise accumulates at each step (``error propagation''), rendering the process highly fragile to precision loss.

\noindent\textbf{(2) Application (KA) is Scale-Responsive.} 
In terms of model size, KA exhibits a high scaling exponent ($\alpha = -0.409$), contrasting with the notably lower exponent of KM ($\alpha = -0.315$). This implies that while memorization capacity saturates faster, application benefits significantly from scaling up, consistent with the ``emergence'' properties often observed in high-level cognitive tasks.

\noindent\textbf{(3) Memorization (KM) is Calibration-Sensitive.} 
L1 tasks show a pronounced sensitivity to calibration data ($\gamma = -0.040$), nearly double that of the more robust KA tasks. We attribute this to KM's reliance on precise activation alignment to trigger Key-Value pairs in FFN layers~\cite{geva_emnlp_2021}. Unlike KA tasks, which rely on generalized patterns robust to numerical shifts, KM's ``exact lookup'' mechanism is susceptible to distribution shifts, necessitating richer calibration data.

\begin{figure*}[t]
    \centering
    \begin{subfigure}[b]{0.32\textwidth}
        \centering
        \includegraphics[width=\textwidth]{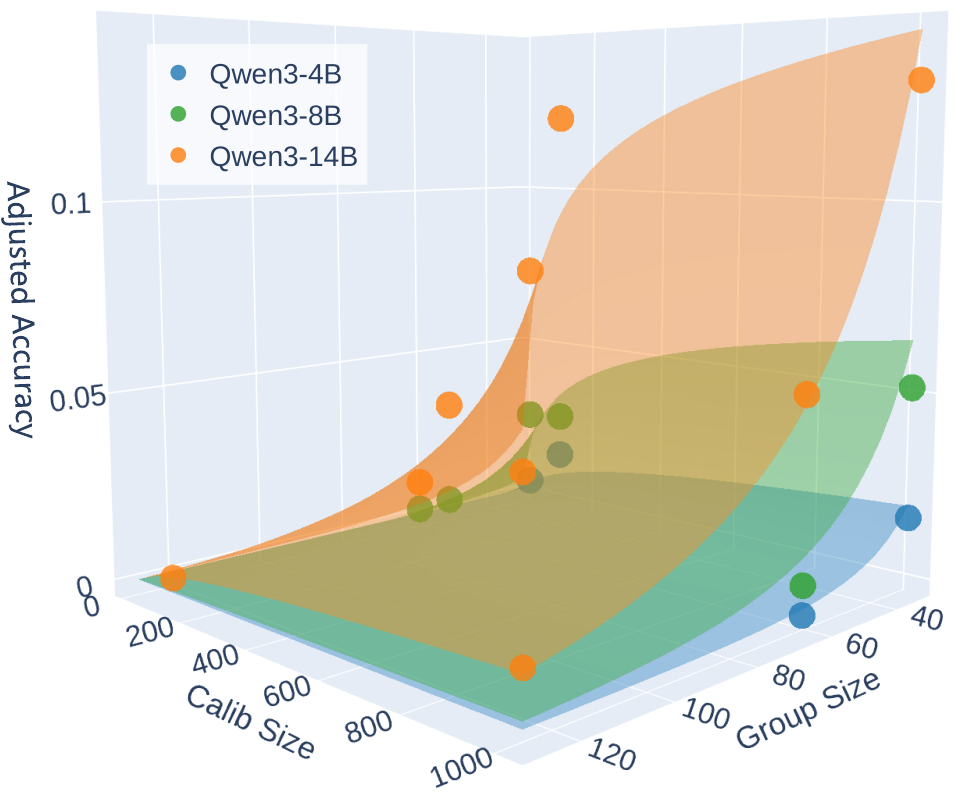}
        \caption{Memorization (KM)}
    \end{subfigure}
    \hfill
    \begin{subfigure}[b]{0.32\textwidth}
        \centering
        \includegraphics[width=\textwidth]{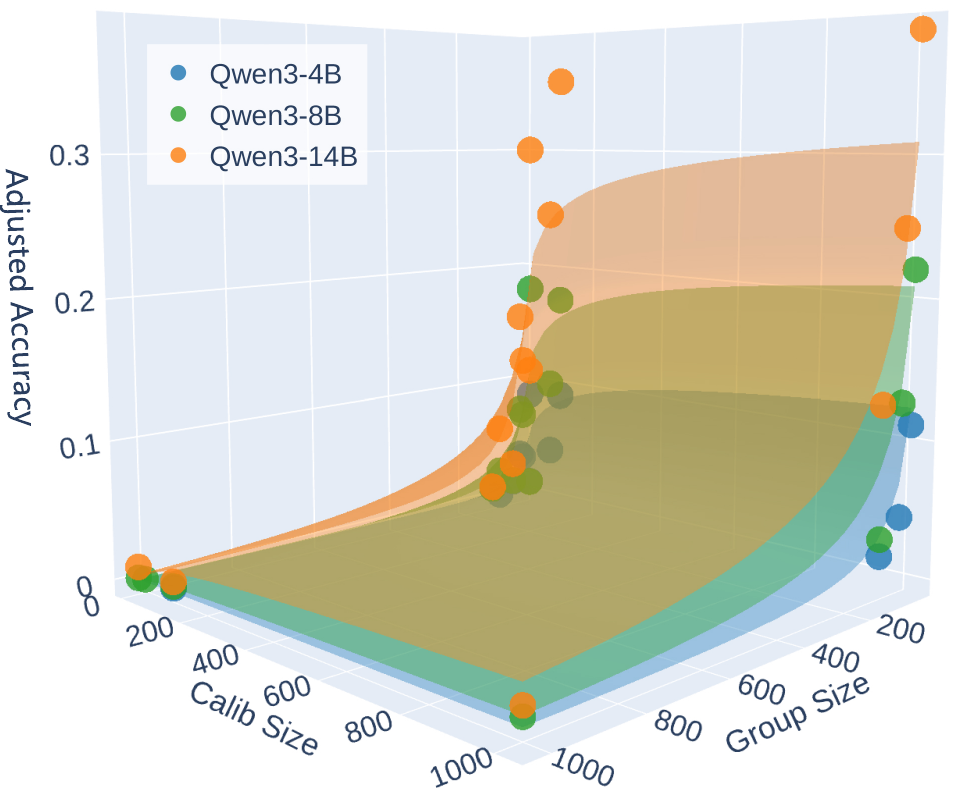}
        \caption{Application (KA)}
    \end{subfigure}
    \hfill
    \begin{subfigure}[b]{0.32\textwidth}
        \centering
        \includegraphics[width=\textwidth]{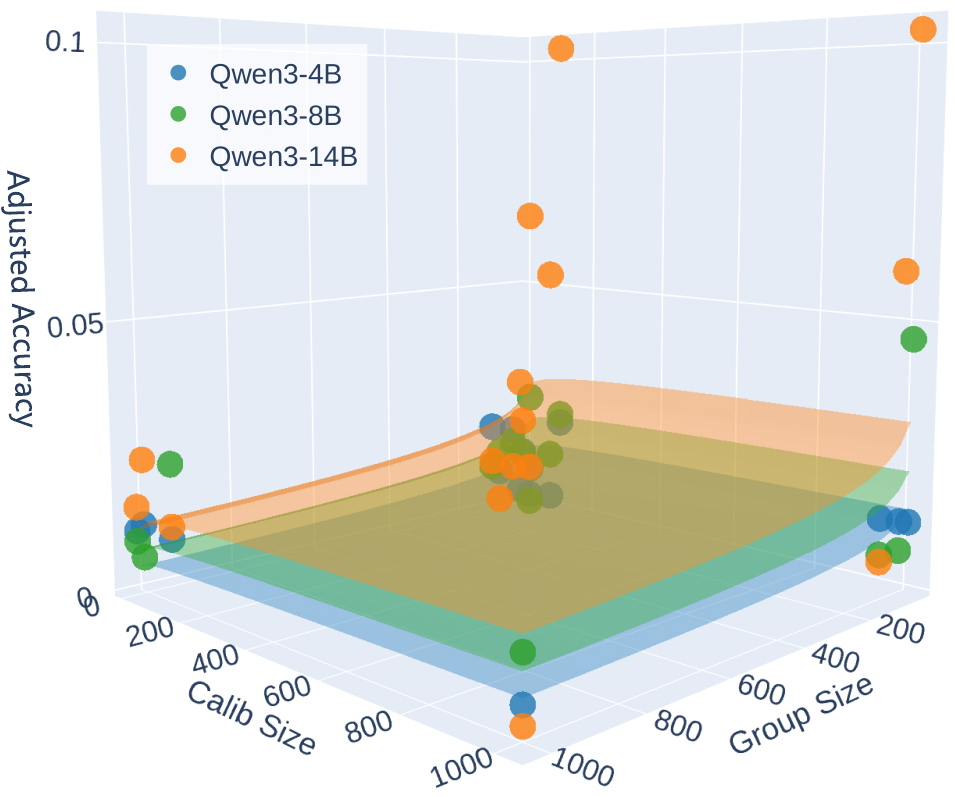}
        \caption{Reasoning (KR)}
    \end{subfigure}
    \caption{
        Fitted performance surfaces under 2-bit quantization ($N \ge 4$B). 
        (a) KM and (b) KA retain robust scaling behaviors with high goodness-of-fit (Adj.$R^2_{\mathcal{O}} \approx 0.91$ and $0.87$, respectively), exhibiting pronounced sensitivity to $G$ and $C_b$. In contrast, (c) KR exhibits a flat surface with poor fit (Adj.$R^2_{\mathcal{O}} \approx 0.22$), indicating a structural collapse of reasoning capabilities regardless of configuration adjustments.
    }
    \label{fig:2bit_surface}
\end{figure*}

\subsubsection{The ``Phase Transition'' at 2-bit}
\label{sec:2bit_phase_transition}

We characterize the entry into the 2-bit region as a critical ``Phase Transition,'' where the scaling behavior diverges sharply depending on model size and task type.

\noindent\textbf{(1) Systemic Collapse in Small-Scale Models.}
For models with $N < 2$B, we observe a universal performance collapse across all tasks. Scaling laws fail to converge (Adj.$R^2_{\mathcal{O}} < 0$). Consequently, PTQ tuning becomes ineffective, as the model lacks the fundamental capacity to retain utility.

\noindent\textbf{(2) Capability Recovery in Large-Scale Models.}
In contrast, larger models ($N \ge 4$B) can maintain capabilities, but with certain conditions. As shown in Figure~\ref{fig:2bit_surface}, while reasoning (KR) fails completely, memorization (KM) and application (KA) are effectively recovered if fine-grained parameters are optimized. Specifically, the scaling exponents for $G$ surges from $\sim 0.07$ (General) to $\sim 0.60$ (KM) and $\sim 0.33$ (KA), and calibration dependence intensifies ($\gamma \approx -0.58$). This implies that using smaller group sizes and sufficient calibration data is no longer optional, but essential for preventing failure in the 2-bit region.


\subsection{Cross-Architecture Validation on Llama-3}

To verify the universality of our framework beyond Qwen, we extend the evaluation to the Llama-3 family (1B, 3B, 8B)~\cite{grattafiori_llama_2024} using consistent quantization strategy and benchmarks. We assess a representative subset of 42 configurations within the effective compression zone.

\noindent\textbf{Universality of the Scaling Framework.}
As shown in Table~\ref{tab:llama3_scaling}, fitting the four-factor formulation yields exceptional goodness-of-fit, with Adj.$R_{\mathcal{O}}^{2}$ exceeding 0.92 across all knowledge levels. This confirms that our multiplicative power-law formulation captures fundamental quantization dynamics independent of architecture. Appendix~\ref{sec:appendix_llama3} provides further visualizations and statistical validation for these results.

\noindent\textbf{Consistency of Knowledge Sensitivities.}
Crucially, the fitted coefficients reinforce the distinct sensitivities observed in Qwen3:
\begin{itemize}[leftmargin=*, topsep=1pt, itemsep=1pt, parsep=1pt]
    \item \emph{Precision Critical}: KR remains the most fragile, showing the highest sensitivity to both bit-width ($\beta$) and group size ($\delta$).
    \item \emph{Scale Responsive}: KA exhibits the highest scaling exponent ($\alpha$) while maintaining the lowest sensitivity to quantization coefficients. This confirms it benefits most from model scaling and is relatively robust to quantization.
    \item \emph{Calibration Sensitive}: Both KM and KR exhibit heightened sensitivity to calibration data compared to the robust KA. This reinforces our finding that while KA is largely scale-driven, retaining memorization and reasoning capabilities necessitates high-quality quantization parameters.
\end{itemize}

\begin{table}[t]
\centering
\footnotesize
\setlength{\tabcolsep}{2.2pt} 
\begin{tabular}{l|c|cccc|c}
\toprule
\textbf{Task} & \textbf{Const}($A$) & $\alpha(N)$ & $\beta(B)$ & $\gamma(C_b)$ & $\delta(G)$ & \textbf{Adj.} $R_{\mathcal{O}}^{2}$ \\
\midrule
\textbf{General} & $2.91\text{e}3$ & -0.333 & -1.501 & -0.056 & 0.072 & 0.9595 \\
\midrule
\textbf{L1: KM} & $5.32\text{e}2$ & -0.249 & -1.596 & -0.060 & 0.074 & 0.9622 \\
\textbf{L2: KA} & $2.19\text{e}4$ & \textbf{-0.447} & -1.462 & -0.045 & 0.073 & 0.9709 \\
\textbf{L3: KR} & $9.66\text{e}3$ & -0.373 & \textbf{-1.645} & \textbf{-0.071} & \textbf{0.080} & 0.9277 \\
\bottomrule
\end{tabular}
\caption{Fitted scaling parameters for Llama-3 family.}
\label{tab:llama3_scaling}
\end{table}


\section{Conclusion}
In this work, we formulate Task-Stratified Knowledge Scaling Laws, integrating model size, bit-width, and crucial fine-grained factors (group size and calibration set size) into a unified framework. 
Validated on 293 diverse configurations, our framework demonstrates strong fit and cross-architecture consistency. 
We identify distinct sensitivities across knowledge capabilities: reasoning is precision-critical, application is scale-responsive, and memorization is calibration-sensitive. Furthermore, we emphasize that under low-bit quantization, optimizing fine-grained factors is essential to prevent performance collapse.

 



\section*{Limitations}
Our study primarily establishes task-stratified PTQ scaling laws for representative dense Transformer architectures under weight-only quantization. While the proposed framework covers diverse knowledge capabilities, future research could extend these laws to other quantization paradigms (e.g., activation quantization) and alternative architectures, such as Mixture-of-Experts (MoE).

\section*{Acknowledgments}
This work was supported by Beijing Natural Science Foundation (L243006), the National Natural Science Foundation of China (No.62406321), the independent research project of the Key Laboratory of Cognition and Decision Intelligence for Complex Systems and CIPS-SMP-Zhipu Large Model Fund.


\bibliography{main}

\appendix

\clearpage

\section{Experimental Details}
\label{sec:appendix_exp}

This appendix provides supplementary details to support the reproducibility of our experiments, covering implementation specifics, benchmark stratification rationale, and full model configurations.

\subsection{Implementation and Evaluation Setup}

\paragraph{Quantization Implementation.} 
Experiments are conducted using the Hugging Face Transformers library~\cite{wolf_transformers_2020}, with GPTQ implemented via the GPTQModel library. We employ default hyperparameters unless otherwise specified. To establish a domain-agnostic baseline, we chose the universal C4 dataset~\cite{raffel_exploring_2020} as our calibration corpus. Samples are randomly drawn with a fixed sequence length of 2048.

\paragraph{Evaluation Framework.} 
We utilize the Language Model Evaluation Harness (\texttt{lm-eval}, v0.4.9) framework~\cite{sutawika_eleutherailm-evaluation-harness_2025} for standardized testing. Most tasks are evaluated in a 5-shot setting. For multiple-choice tasks, we report the ``acc\_norm'' (accuracy normalized by choice length) to mitigate length bias, while generative tasks use ``exact\_match''. A specific exception is the TREx benchmark (part of LAMA). We strictly control the prompt variance: for each of the 39 relation types, we select the single prompt template from the Pararel dataset~\cite{elazar_measuring_2021} where the object [Y] is positioned at the end of the sentence. Performance for TREx is reported using the Precision@5 (P@5) metric.

\subsection{Benchmarks Mapping and Statistics}
\label{sub:bench_mapping}

Table~\ref{tab:dataset_stats_detailed} provides a comprehensive mapping of the 14 benchmarks to our cognitive taxonomy, along with their statistical details.

\begin{table*}[!t]
\centering
\small
\renewcommand{\arraystretch}{1.2} 
\setlength{\tabcolsep}{5pt} 
\resizebox{\textwidth}{!}{%
\begin{tabular}{llllcccl}
\toprule
\textbf{Level} & \textbf{Benchmark} & \textbf{Domain} & \textbf{Type} & \textbf{Metric} & \textbf{Baseline} & \textbf{Size} & \textbf{Characteristics} \\
\midrule
\multirow{5}{*}{\textbf{KM}} 
 & TriviaQA & Trivia \& Web & Gen & EM & $\approx 0$ & 17,944 & \multirow{5}{*}{\parbox{5cm}{\textbf{Factual Recall}: Requires recalling precise entities (names, dates) from the internal knowledge base without complex contextual transformation.}} \\
 & Natural Questions & Wikipedia & Gen & EM & $\approx 0$ & 3,610 & \\
 & WebQuestions & KB (Freebase) & Gen & EM & $\approx 0$ & 2,032 & \\
 & TREx (LAMA) & KB (Wikidata) & Gen & P@5 & $\approx 0$ & 27,610 & \\
 & SQuAD (LAMA) & Wikipedia & Gen & P@5 & $\approx 0$ & 212 & \\
\midrule
\multirow{4}{*}{\textbf{KA}} 
 & MMLU & 57 Subjects & MC (4) & Acc & 0.25 & 14,042 & \multirow{4}{*}{\parbox{5cm}{\textbf{Flexible Application}: Requires understanding contexts and applying internalized knowledge for specific scenarios, emphasizing flexible utilization rather than strict facts lookup.}} \\
 & Hellaswag & Commonsense & MC (4) & Acc & 0.25 & 10,042 & \\
 & Winogrande & Commonsense & MC (2) & Acc & 0.50 & 1,267 & \\
 & ARC-Easy & Science (Basic) & MC (4) & Acc & 0.25 & 2,376 & \\
\midrule
\multirow{5}{*}{\textbf{KR}} 
 & ARC-Challenge & Science (Hard) & MC (4) & Acc & 0.25 & 1,172 & \multirow{5}{*}{\parbox{5cm}{\textbf{Multi-step Reasoning}: Requires constructing sequential logical chains (mathematical derivation or multi-hop logic).}} \\
 & StrategyQA & Open-Domain & MC (2) & Acc & 0.50 & 2,289 & \\
 & OpenbookQA & Science \& Common & MC (4) & Acc & 0.25 & 500 & \\
 & MathQA & Math & MC (5) & Acc & 0.20 & 2,985 & \\
 & GSM8K & Math & Gen & EM & $\approx 0$ & 1,319 & \\
\bottomrule
\end{tabular}%
}
\caption{Detailed statistics and cognitive mapping of benchmarks. \textbf{Type} denotes the task format (Generative vs. Multiple-Choice). \textbf{Metric} denotes Exact Match (EM), Accuracy (Acc), or Precision@5 (P@5). \textbf{Characteristics} justifies the classification by highlighting the underlying task nature.}
\label{tab:dataset_stats_detailed}
\end{table*}

\begin{table*}[!t]
  \centering
  \small
  \renewcommand{\arraystretch}{1.5}
  \setlength{\tabcolsep}{6pt}
  \begin{tabular*}{\textwidth}{@{\extracolsep{\fill}} l l c c @{}}
    \toprule
    \textbf{Formulation} & \textbf{Fitted Function} & \textbf{Adj. $R^2_{\mathcal{L}}$} & \textbf{Adj. $R^2_{\mathcal{O}}$} \\
    \midrule
    \multicolumn{4}{l}{\textit{\textbf{L1: Knowledge Memorization (KM)}}} \\
    $f(N, B, C_b, G)$ & $2.08 \times 10^3 \, N^{-0.315} \, (\log_2 B)^{-0.964} \, (\log_2 C_b)^{-0.040} \, G^{0.064}$ & \textbf{0.9341} & \textbf{0.9350} \\
    $f(N, B)$       & $2.51 \times 10^3 \, N^{-0.313} \, (\log_2 B)^{-0.959}$ & 0.8946 & 0.8993 \\
    $f(N, B, G)$    & $1.95 \times 10^3 \, N^{-0.315} \, (\log_2 B)^{-0.969} \, G^{0.064}$ & 0.9328 & 0.9326 \\
    $f(N, B, C_b)$  & $2.69 \times 10^3 \, N^{-0.313} \, (\log_2 B)^{-0.953} \, (\log_2 C_b)^{-0.040}$ & 0.8957 & 0.9015 \\
    
    \midrule
    \multicolumn{4}{l}{\textit{\textbf{L2: Knowledge Application (KA)}}} \\
    $f(N, B, C_b, G)$ & $7.37 \times 10^3 \, N^{-0.409} \, (\log_2 B)^{-0.982} \, (\log_2 C_b)^{-0.023} \, G^{0.069}$ & \textbf{0.9550} & \textbf{0.9626} \\
    $f(N, B)$       & $9.89 \times 10^3 \, N^{-0.409} \, (\log_2 B)^{-0.986}$ & 0.9276 & 0.9362 \\
    $f(N, B, G)$    & $7.09 \times 10^3 \, N^{-0.409} \, (\log_2 B)^{-0.986} \, G^{0.069}$ & 0.9549 & 0.9624 \\
    $f(N, B, C_b)$  & $1.03 \times 10^4 \, N^{-0.409} \, (\log_2 B)^{-0.982} \, (\log_2 C_b)^{-0.023}$ & 0.9275 & 0.9362 \\
    
    \midrule
    \multicolumn{4}{l}{\textit{\textbf{L3: Knowledge Reasoning (KR)}}} \\
    $f(N, B, C_b, G)$ & $1.27 \times 10^4 \, N^{-0.405} \, (\log_2 B)^{-1.356} \, (\log_2 C_b)^{-0.034} \, G^{0.087}$ & \textbf{0.9156} & \textbf{0.9218} \\
    $f(N, B)$       & $1.55 \times 10^4 \, N^{-0.398} \, (\log_2 B)^{-1.330}$ & 0.8738 & 0.8775 \\
    $f(N, B, G)$    & $1.20 \times 10^4 \, N^{-0.405} \, (\log_2 B)^{-1.361} \, G^{0.087}$ & 0.9154 & 0.9212 \\
    $f(N, B, C_b)$  & $1.64 \times 10^4 \, N^{-0.398} \, (\log_2 B)^{-1.325} \, (\log_2 C_b)^{-0.034}$ & 0.8738 & 0.8779 \\
    \bottomrule
  \end{tabular*}
  \caption{Ablation analysis for task-stratified scaling laws across three knowledge levels. Including fine-grained factors ($G, C_b$) consistently improves goodness-of-fit. The model form is $-\ln(\mathrm{Acc}_{\text{adj}}) = A \cdot N^{\alpha} (\log_2 B)^{\beta} (\log_2 C_b)^{\gamma} G^{\delta}$.}
  \label{tab:app_ablation}
\end{table*}

\subsection{Full Experimental Configurations}
\label{sub:full_configs}

To ensure reproducibility and transparency, Table~\ref{tab:full_configs} enumerates all 293 experimental configurations evaluated in this study, covering the Main (scaling fit), Validation, and Generalization groups.

\subsection{Numerical Stabilization in Regression}
\label{subsec:numerical_stab}

To ensure numerical stability during regression, we implement filtering rules for the transformation $\ln(-\ln(\mathrm{Acc}_{\text{adj}}))$. Because this term is undefined for $\mathrm{Acc}_{\text{adj}} \leq 0$ and approaches a mathematical singularity as $\mathrm{Acc}_{\text{adj}} \to 0^+$, we establish a lower-bound threshold of $\epsilon = 0.01$. Configurations yielding $\mathrm{Acc}_{\text{adj}} \leq 0.01$ are considered ``collapsed to random guessing'' and are excluded. At this boundary, the transformation yields $\ln(-\ln(0.01)) \approx 1.527$, ensuring stable computation.

In our main experiments on the Qwen3 family, exactly 6 configurations trigger this filter. All of them share the most aggressive compression setting: the smallest model size ($N=0.6\mathrm{B}$) at 3-bit weight precision with the coarsest group size ($G=1024$).

\section{Definition of Adjusted $R^2$}
\label{appendix:r2}

While the standard coefficient of determination ($R^2$) measures the proportion of variance explained by the model, it tends to increase when more variables are added, regardless of their actual predictive power. To provide a robust assessment that accounts for model complexity, we employ the Adj. $R^2$ (denoted as $R^2_{adj}$).

First, the standard $R^2$ is defined as:
\begin{equation}
R^2 = 1 - \frac{\sum_{i=1}^{n} (y_i - \hat{y}_i)^2}{\sum_{i=1}^{n} (y_i - \bar{y})^2} \,,
\end{equation}
where \( y_i \) is the true value, \( \hat{y}_i \) is the model prediction, and \( \bar{y} \) is the empirical mean of the true values.

The Adj. $R^2$ is then calculated as:
\begin{equation}
R^2_{adj} = 1 - (1 - R^2) \frac{n - 1}{n - p - 1} \,,
\end{equation}
where $n$ is the sample size (number of observations) and $p$ is the number of predictors (independent variables) in the fitted model. Unlike standard $R^2$, the Adj. $R^2$ penalizes the inclusion of non-informative parameters, ensuring that the reported goodness-of-fit accurately reflects the model's explanatory power relative to its complexity.

\section{Detailed Ablation and Statistical Significance for Qwen3}
\label{sec:appendix_qwen3}

\subsection{Ablation Study on Fine-Grained Factors}
\label{subsec:ablation_qwen3}

To validate the necessity of including Group Size ($G$) and Calibration Set Size ($C_b$) in our task-stratified scaling laws, we conduct ablation studies across the three knowledge capabilities (KM, KA, KR), as detailed in Table~\ref{tab:app_ablation}.

The results consistently highlight two patterns. First, adding $G$ significantly enhances the goodness-of-fit across all tasks (e.g., KR improves from $0.8775 \to 0.9212$), confirming quantization granularity as a universal determinant. Second, the impact of $C_b$ varies by task nature: it yields negligible improvement for the robust Knowledge Application (KA) task, but provides detectable gains for Knowledge Memorization (KM) and Reasoning (KR). This empirical evidence reinforces the sensitivity hierarchy discussed in Section~\ref{sec:stratified}, where specific capabilities rely more heavily on precise distribution alignment.

\clearpage

\begin{table}[!t]
    \centering
    \small
    \setlength{\tabcolsep}{11pt} 
    \renewcommand{\arraystretch}{1.2}
    \begin{tabular}{@{}lccc@{}} 
        \toprule
         & \multicolumn{2}{c}{\textbf{Qwen3}} & \textbf{Llama-3} \\
        \cmidrule(lr){2-3} \cmidrule(l){4-4}
        \textbf{Task Level} & \textbf{Fit} & \textbf{Validation} & \textbf{Fit} \\
        \midrule
        \textbf{General} & 0.0267 & 0.0630 & 0.0171 \\
        \textbf{L1: KM}  & 0.0219 & 0.0946 & 0.0147 \\
        \textbf{L2: KA}  & 0.0254 & 0.0292 & 0.0172 \\
        \textbf{L3: KR}  & 0.0361 & 0.0555 & 0.0227 \\
        \bottomrule
    \end{tabular}
    \caption{Mean Absolute Error (MAE) of the predicted accuracy. `Validation' denotes the held-out Qwen3-32B.}
    \label{tab:mae_results}
\end{table}

\subsection{Statistical Significance of Scaling Exponents}
\label{subsec:stats_qwen3}

To rigorously validate that the observed sensitivities in the Qwen3 family are not artifacts of random variance, we compute the Standard Errors (SE) and 95\% Confidence Intervals (CI) for all fitted exponents ($\alpha, \beta, \gamma, \delta$). The regressions were evaluated using the \texttt{statsmodels} library. An exponent is considered statistically significant if its 95\% CI strictly excludes zero.

The results, presented in Table~\ref{tab:stats_qwen}, align with our qualitative observations. Across all task levels, the primary drivers ($N, B, G$) are highly significant. Notably, the coefficient for calibration set size ($\gamma(C_b)$) is strictly negative for KM (e.g., $[-0.078, -0.002]$), confirming its calibration-sensitive nature, whereas it crosses zero for KA ($[-0.063, +0.016]$), indicating that application capabilities rely fundamentally on model scale rather than granular calibration alignment.

\begin{table*}[!t]
  \centering
  \small
  \renewcommand{\arraystretch}{1.2}
  \setlength{\tabcolsep}{6pt}
  \begin{tabular}{llcccc}
    \toprule
    \textbf{Task Level} & \textbf{Metric} & \textbf{$\alpha(N)$} & \textbf{$\beta(B)$} & \textbf{$\gamma(C_b)$} & \textbf{$\delta(G)$} \\
    \midrule
    \multirow{2}{*}{\textbf{General}} 
    & Est. $\pm$ SE & $-0.359 \pm 0.008$ & $-1.067 \pm 0.062$ & $-0.033 \pm 0.021$ & $+0.074 \pm 0.007$ \\
    & 95\% CI & $[-0.374, -0.344]$ & $[-1.189, -0.944]$ & $[-0.073, +0.008]$ & $[+0.060, +0.087]$ \\
    \cmidrule{1-6}
    \multirow{2}{*}{\textbf{L1: KM}} 
    & Est. $\pm$ SE & $-0.315 \pm 0.007$ & $-0.964 \pm 0.058$ & $\mathbf{-0.040 \pm 0.019}$ & $+0.065 \pm 0.007$ \\
    & 95\% CI & $[-0.329, -0.301]$ & $[-1.078, -0.849]$ & $\mathbf{[-0.078, -0.002]}$ & $[+0.051, +0.078]$ \\
    \cmidrule{1-6}
    \multirow{2}{*}{\textbf{L2: KA}} 
    & Est. $\pm$ SE & $\mathbf{-0.409 \pm 0.007}$ & $-0.982 \pm 0.061$ & $-0.023 \pm 0.020$ & $+0.069 \pm 0.007$ \\
    & 95\% CI & $\mathbf{[-0.424, -0.395]}$ & $[-1.102, -0.863]$ & $[-0.063, +0.016]$ & $[+0.055, +0.082]$ \\
    \cmidrule{1-6}
    \multirow{2}{*}{\textbf{L3: KR}} 
    & Est. $\pm$ SE & $-0.405 \pm 0.011$ & $\mathbf{-1.356 \pm 0.085}$ & $-0.034 \pm 0.028$ & $\mathbf{+0.087 \pm 0.010}$ \\
    & 95\% CI & $[-0.425, -0.384]$ & $\mathbf{[-1.523, -1.189]}$ & $[-0.089, +0.022]$ & $\mathbf{[+0.068, +0.107]}$ \\
    \bottomrule
  \end{tabular}
  \caption{Statistical significance of the fitted scaling exponents for the Qwen3 family. CIs that strictly exclude zero indicate statistical significance.}
  \label{tab:stats_qwen}
\end{table*}


\begin{figure*}[!t]
    \centering
    \begin{subfigure}[b]{0.32\textwidth}
        \centering
        \includegraphics[width=\textwidth]{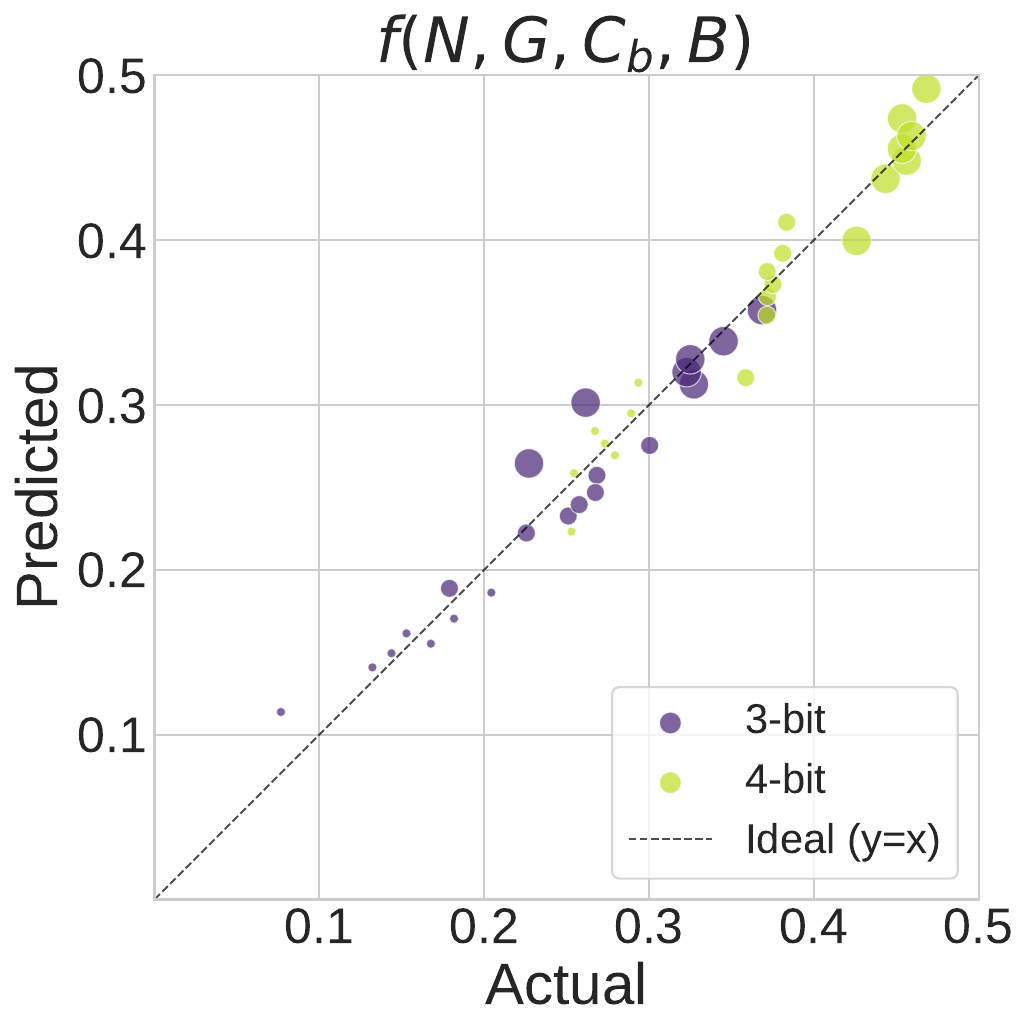}
        \caption{Memorization (KM)}
        \label{fig:llama3_km}
    \end{subfigure}
    \hfill
    \begin{subfigure}[b]{0.32\textwidth}
        \centering
        \includegraphics[width=\textwidth]{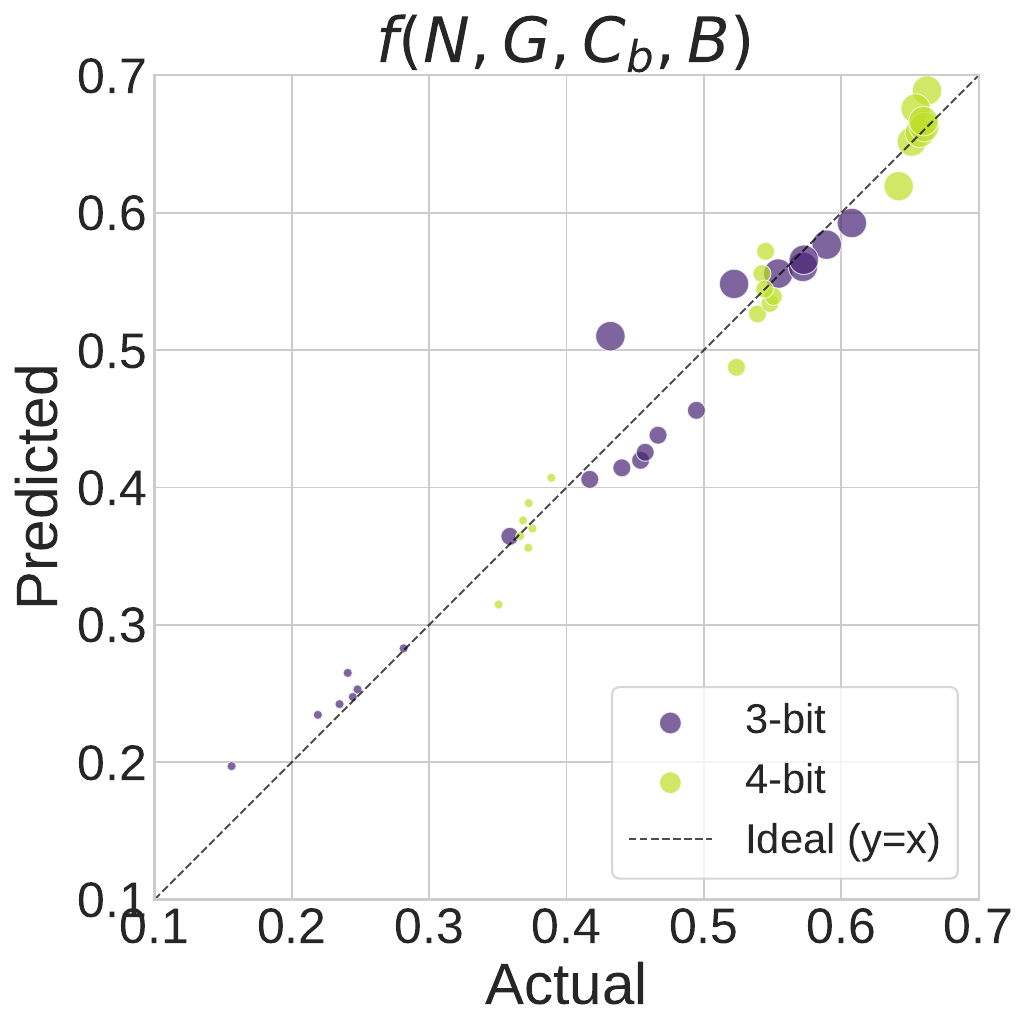}
        \caption{Application (KA)}
        \label{fig:llama3_ka}
    \end{subfigure}
    \hfill
    \begin{subfigure}[b]{0.32\textwidth}
        \centering
        \includegraphics[width=\textwidth]{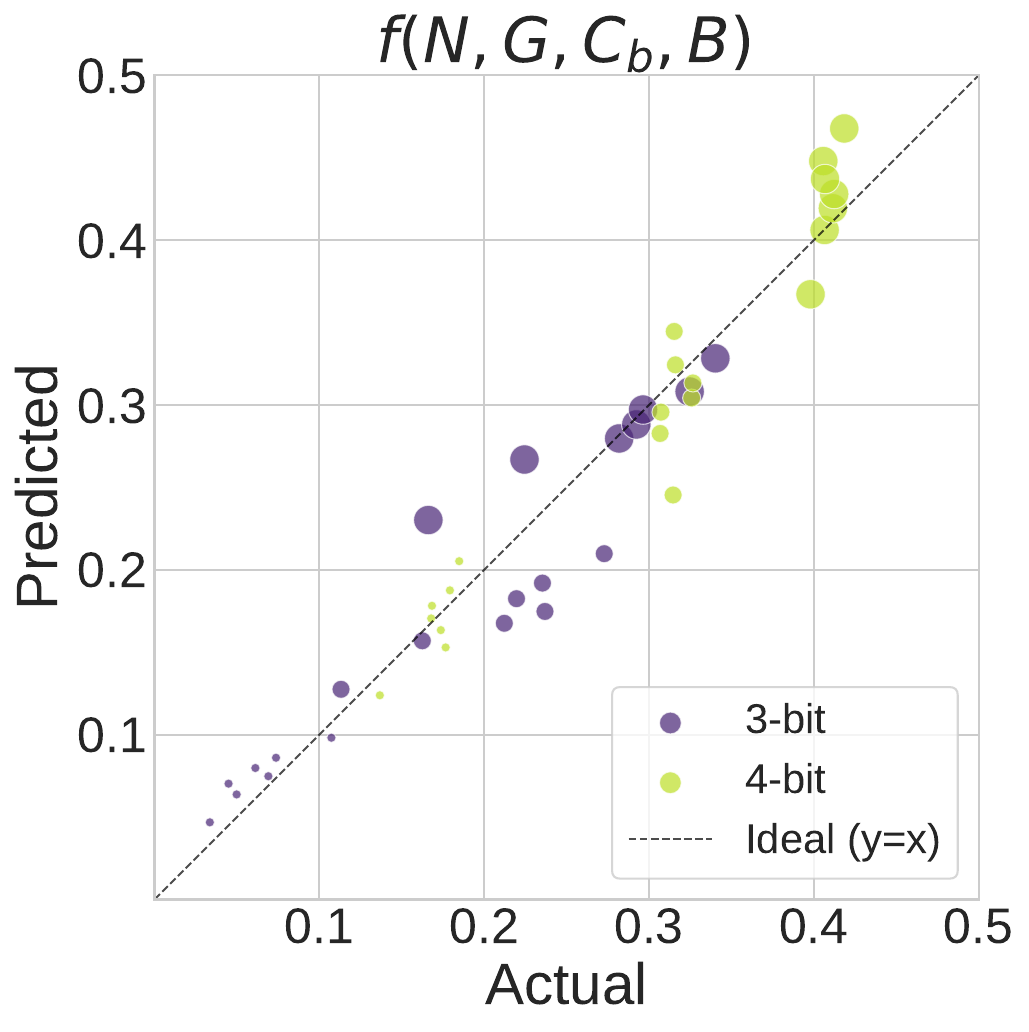}
        \caption{Reasoning (KR)}
        \label{fig:llama3_kr}
    \end{subfigure}
    \caption{Goodness-of-fit visualization for Llama-3 family. The scatter plots compare the predicted adjusted accuracy (y-axis) against the actual empirical values (x-axis) for (a) Memorization, (b) Application, and (c) Reasoning. The close alignment with the dashed diagonal line ($y=x$) indicates high predictive accuracy. Point size corresponds to model size (1B, 3B, 8B), and color indicates bit-width.}
    \label{fig:llama3_scatter}
\end{figure*}

\section{Cross-Architecture Validation and Predictive Robustness}
\label{sec:appendix_cross_arch}

\subsection{Scaling Law Analysis on Llama-3}
\label{sec:appendix_llama3}

\textbf{Experimental Configuration Details.}
For the Llama-3 generalization experiments, we analyze 42 representative quantization configurations in the effective compression zone (3-bit and 4-bit). To efficiently traverse the hyperparameter space, we adopt a controlled grid search strategy: (1) Fixed Group Size ($G=128$) with varying Calibration Set Sizes ($C_b \in \{8, 32, 128, 1024\}$); and (2) Fixed Calibration Set Size ($C_b=128$) with varying Group Sizes ($G \in \{32, 64, 128, 1024\}$). This setup ensures coverage of key sensitivity thresholds while maintaining computational feasibility.

\vspace{0.5em}
\noindent\textbf{Visualization.}
Figure~\ref{fig:llama3_scatter} compares the predicted and actual adjusted accuracy across all three knowledge levels for the Llama-3 family, visually confirming the high goodness-of-fit reported in the main text.


\vspace{0.5em}
\noindent\textbf{Statistical Significance.} As shown in Table~\ref{tab:stats_llama3}, the statistical behaviors strictly mirror those of Qwen3. The primary factors remain highly significant (95\% CIs exclude zero). Crucially, the unique sensitivity of Knowledge Memorization to calibration is preserved ($\gamma(C_b)$ CI is $[-0.113, -0.007]$), confirming that this calibration dependence is an architecture-agnostic property of memorization tasks.

\begin{table*}[t]
  \centering
  \small
  \renewcommand{\arraystretch}{1.2}
  \setlength{\tabcolsep}{6pt}
  \begin{tabular}{llcccc}
    \toprule
    \textbf{Task Level} & \textbf{Metric} & \textbf{$\alpha(N)$} & \textbf{$\beta(B)$} & \textbf{$\gamma(C_b)$} & \textbf{$\delta(G)$} \\
    \midrule
    \multirow{2}{*}{\textbf{General}} 
    & Est. $\pm$ SE & $-0.333 \pm 0.013$ & $-1.501 \pm 0.091$ & $-0.056 \pm 0.030$ & $+0.072 \pm 0.011$ \\
    & 95\% CI & $[-0.360, -0.307]$ & $[-1.685, -1.318]$ & $[-0.117, +0.005]$ & $[+0.050, +0.093]$ \\
    \cmidrule{1-6}
    \multirow{2}{*}{\textbf{L1: KM}} 
    & Est. $\pm$ SE & $-0.249 \pm 0.011$ & $-1.596 \pm 0.079$ & $\mathbf{-0.060 \pm 0.026}$ & $+0.074 \pm 0.009$ \\
    & 95\% CI & $[-0.273, -0.226]$ & $[-1.756, -1.437]$ & $\mathbf{[-0.113, -0.007]}$ & $[+0.055, +0.093]$ \\
    \cmidrule{1-6}
    \multirow{2}{*}{\textbf{L2: KA}} 
    & Est. $\pm$ SE & $\mathbf{-0.447 \pm 0.014}$ & $-1.462 \pm 0.095$ & $-0.045 \pm 0.032$ & $+0.073 \pm 0.011$ \\
    & 95\% CI & $\mathbf{[-0.475, -0.419]}$ & $[-1.655, -1.268]$ & $[-0.109, +0.019]$ & $[+0.050, +0.096]$ \\
    \cmidrule{1-6}
    \multirow{2}{*}{\textbf{L3: KR}} 
    & Est. $\pm$ SE & $-0.373 \pm 0.019$ & $\mathbf{-1.645 \pm 0.131}$ & $-0.071 \pm 0.044$ & $\mathbf{+0.080 \pm 0.016}$ \\
    & 95\% CI & $[-0.411, -0.334]$ & $\mathbf{[-1.911, -1.379]}$ & $[-0.159, +0.018]$ & $\mathbf{[+0.048, +0.111]}$ \\
    \bottomrule
  \end{tabular}
  \caption{Statistical significance of the fitted scaling exponents for the Llama-3 family, demonstrating cross-architecture consistency.}
  \label{tab:stats_llama3}
\end{table*}

\Needspace{10\baselineskip}
\subsection{Predictive Quality and Extrapolation}
\label{subsec:mae_results}

While the Adjusted $R^2$ measures the explained variance, deployment decisions often require evaluating the absolute predictive error. To this end, we compute the Mean Absolute Error (MAE) across both the Qwen3 and Llama-3 families. 

As shown in Table~\ref{tab:mae_results}, the MAE remains remarkably low. Crucially, scale-extrapolation on the held-out Qwen3-32B validation model is highly reliable, with prediction errors for KA and KR strictly bounded to $\approx 2.9\%$ and $\approx 5.5\%$ respectively.


\clearpage

\onecolumn 

\begin{small} 
\setlength{\tabcolsep}{3.0pt} 
\renewcommand{\arraystretch}{1.1} 
\begin{longtable}{c c c c c c c | c c c c c c c}
\caption{All configurations of experiments.
The Type column classifies the 293 data points into three roles: Fit (245 Qwen3 configurations for fitting scaling coefficients), Val (6 held-out Qwen3-32B configurations for extrapolation validation), and Gen (42 Llama-3 configurations for cross-architecture generalization).
}
\label{tab:full_configs} \\

\toprule
\textbf{No.} & \textbf{Model} & \textbf{$N$} & \textbf{$B$} & \textbf{$G$} & \textbf{$C_b$} & \textbf{Type} & 
\textbf{No.} & \textbf{Model} & \textbf{$N$} & \textbf{$B$} & \textbf{$G$} & \textbf{$C_b$} & \textbf{Type} \\
\midrule
\endfirsthead

\multicolumn{14}{c}{{\bfseries \tablename\ \thetable{} -- continued from previous page}} \\
\toprule
\textbf{No.} & \textbf{Model} & \textbf{$N$} & \textbf{$B$} & \textbf{$G$} & \textbf{$C_b$} & \textbf{Type} & 
\textbf{No.} & \textbf{Model} & \textbf{$N$} & \textbf{$B$} & \textbf{$G$} & \textbf{$C_b$} & \textbf{Type} \\
\midrule
\endhead

\midrule
\multicolumn{14}{r}{{Continued on next page}} \\
\endfoot

\bottomrule
\endlastfoot

0 & Qwen3-0.6B & 440,467,456 & 8 & 128 & 128 & Fit & 1 & Qwen3-0.6B & 440,467,456 & 4 & 32 & 8 & Fit \\
2 & Qwen3-0.6B & 440,467,456 & 4 & 32 & 32 & Fit & 3 & Qwen3-0.6B & 440,467,456 & 4 & 32 & 128 & Fit \\
4 & Qwen3-0.6B & 440,467,456 & 4 & 32 & 1024 & Fit & 5 & Qwen3-0.6B & 440,467,456 & 4 & 64 & 8 & Fit \\
6 & Qwen3-0.6B & 440,467,456 & 4 & 64 & 32 & Fit & 7 & Qwen3-0.6B & 440,467,456 & 4 & 64 & 128 & Fit \\
8 & Qwen3-0.6B & 440,467,456 & 4 & 64 & 1024 & Fit & 9 & Qwen3-0.6B & 440,467,456 & 4 & 128 & 8 & Fit \\
10 & Qwen3-0.6B & 440,467,456 & 4 & 128 & 32 & Fit & 11 & Qwen3-0.6B & 440,467,456 & 4 & 128 & 128 & Fit \\
12 & Qwen3-0.6B & 440,467,456 & 4 & 128 & 1024 & Fit & 13 & Qwen3-0.6B & 440,467,456 & 4 & 1024 & 8 & Fit \\
14 & Qwen3-0.6B & 440,467,456 & 4 & 1024 & 32 & Fit & 15 & Qwen3-0.6B & 440,467,456 & 4 & 1024 & 128 & Fit \\
16 & Qwen3-0.6B & 440,467,456 & 4 & 1024 & 1024 & Fit & 17 & Qwen3-0.6B & 440,467,456 & 3 & 32 & 8 & Fit \\
18 & Qwen3-0.6B & 440,467,456 & 3 & 32 & 32 & Fit & 19 & Qwen3-0.6B & 440,467,456 & 3 & 32 & 128 & Fit \\
20 & Qwen3-0.6B & 440,467,456 & 3 & 32 & 1024 & Fit & 21 & Qwen3-0.6B & 440,467,456 & 3 & 64 & 8 & Fit \\
22 & Qwen3-0.6B & 440,467,456 & 3 & 64 & 32 & Fit & 23 & Qwen3-0.6B & 440,467,456 & 3 & 64 & 128 & Fit \\
24 & Qwen3-0.6B & 440,467,456 & 3 & 64 & 1024 & Fit & 25 & Qwen3-0.6B & 440,467,456 & 3 & 128 & 8 & Fit \\
26 & Qwen3-0.6B & 440,467,456 & 3 & 128 & 32 & Fit & 27 & Qwen3-0.6B & 440,467,456 & 3 & 128 & 128 & Fit \\
28 & Qwen3-0.6B & 440,467,456 & 3 & 128 & 1024 & Fit & 29 & Qwen3-0.6B & 440,467,456 & 3 & 1024 & 8 & Fit \\
30 & Qwen3-0.6B & 440,467,456 & 3 & 1024 & 32 & Fit & 31 & Qwen3-0.6B & 440,467,456 & 3 & 1024 & 128 & Fit \\
32 & Qwen3-0.6B & 440,467,456 & 3 & 1024 & 1024 & Fit & 33 & Qwen3-0.6B & 440,467,456 & 2 & 32 & 8 & Fit \\
34 & Qwen3-0.6B & 440,467,456 & 2 & 32 & 32 & Fit & 35 & Qwen3-0.6B & 440,467,456 & 2 & 32 & 128 & Fit \\
36 & Qwen3-0.6B & 440,467,456 & 2 & 32 & 1024 & Fit & 37 & Qwen3-0.6B & 440,467,456 & 2 & 64 & 8 & Fit \\
38 & Qwen3-0.6B & 440,467,456 & 2 & 64 & 32 & Fit & 39 & Qwen3-0.6B & 440,467,456 & 2 & 64 & 128 & Fit \\
40 & Qwen3-0.6B & 440,467,456 & 2 & 64 & 1024 & Fit & 41 & Qwen3-0.6B & 440,467,456 & 2 & 128 & 8 & Fit \\
42 & Qwen3-0.6B & 440,467,456 & 2 & 128 & 32 & Fit & 43 & Qwen3-0.6B & 440,467,456 & 2 & 128 & 128 & Fit \\
44 & Qwen3-0.6B & 440,467,456 & 2 & 128 & 1024 & Fit & 45 & Qwen3-0.6B & 440,467,456 & 2 & 1024 & 8 & Fit \\
46 & Qwen3-0.6B & 440,467,456 & 2 & 1024 & 32 & Fit & 47 & Qwen3-0.6B & 440,467,456 & 2 & 1024 & 128 & Fit \\
48 & Qwen3-0.6B & 440,467,456 & 2 & 1024 & 1024 & Fit & 49 & Qwen3-1.7B & 1,409,410,048 & 8 & 128 & 128 & Fit \\
50 & Qwen3-1.7B & 1,409,410,048 & 4 & 32 & 8 & Fit & 51 & Qwen3-1.7B & 1,409,410,048 & 4 & 32 & 32 & Fit \\
52 & Qwen3-1.7B & 1,409,410,048 & 4 & 32 & 128 & Fit & 53 & Qwen3-1.7B & 1,409,410,048 & 4 & 32 & 1024 & Fit \\
54 & Qwen3-1.7B & 1,409,410,048 & 4 & 64 & 8 & Fit & 55 & Qwen3-1.7B & 1,409,410,048 & 4 & 64 & 32 & Fit \\
56 & Qwen3-1.7B & 1,409,410,048 & 4 & 64 & 128 & Fit & 57 & Qwen3-1.7B & 1,409,410,048 & 4 & 64 & 1024 & Fit \\
58 & Qwen3-1.7B & 1,409,410,048 & 4 & 128 & 8 & Fit & 59 & Qwen3-1.7B & 1,409,410,048 & 4 & 128 & 32 & Fit \\
60 & Qwen3-1.7B & 1,409,410,048 & 4 & 128 & 128 & Fit & 61 & Qwen3-1.7B & 1,409,410,048 & 4 & 128 & 1024 & Fit \\
62 & Qwen3-1.7B & 1,409,410,048 & 4 & 1024 & 8 & Fit & 63 & Qwen3-1.7B & 1,409,410,048 & 4 & 1024 & 32 & Fit \\
64 & Qwen3-1.7B & 1,409,410,048 & 4 & 1024 & 128 & Fit & 65 & Qwen3-1.7B & 1,409,410,048 & 4 & 1024 & 1024 & Fit \\
66 & Qwen3-1.7B & 1,409,410,048 & 3 & 32 & 8 & Fit & 67 & Qwen3-1.7B & 1,409,410,048 & 3 & 32 & 32 & Fit \\
68 & Qwen3-1.7B & 1,409,410,048 & 3 & 32 & 128 & Fit & 69 & Qwen3-1.7B & 1,409,410,048 & 3 & 32 & 1024 & Fit \\
70 & Qwen3-1.7B & 1,409,410,048 & 3 & 64 & 8 & Fit & 71 & Qwen3-1.7B & 1,409,410,048 & 3 & 64 & 32 & Fit \\
72 & Qwen3-1.7B & 1,409,410,048 & 3 & 64 & 128 & Fit & 73 & Qwen3-1.7B & 1,409,410,048 & 3 & 64 & 1024 & Fit \\
74 & Qwen3-1.7B & 1,409,410,048 & 3 & 128 & 8 & Fit & 75 & Qwen3-1.7B & 1,409,410,048 & 3 & 128 & 32 & Fit \\
76 & Qwen3-1.7B & 1,409,410,048 & 3 & 128 & 128 & Fit & 77 & Qwen3-1.7B & 1,409,410,048 & 3 & 128 & 1024 & Fit \\
78 & Qwen3-1.7B & 1,409,410,048 & 3 & 1024 & 8 & Fit & 79 & Qwen3-1.7B & 1,409,410,048 & 3 & 1024 & 32 & Fit \\
80 & Qwen3-1.7B & 1,409,410,048 & 3 & 1024 & 128 & Fit & 81 & Qwen3-1.7B & 1,409,410,048 & 3 & 1024 & 1024 & Fit \\
82 & Qwen3-1.7B & 1,409,410,048 & 2 & 32 & 8 & Fit & 83 & Qwen3-1.7B & 1,409,410,048 & 2 & 32 & 32 & Fit \\
84 & Qwen3-1.7B & 1,409,410,048 & 2 & 32 & 128 & Fit & 85 & Qwen3-1.7B & 1,409,410,048 & 2 & 32 & 1024 & Fit \\
86 & Qwen3-1.7B & 1,409,410,048 & 2 & 64 & 8 & Fit & 87 & Qwen3-1.7B & 1,409,410,048 & 2 & 64 & 32 & Fit \\
88 & Qwen3-1.7B & 1,409,410,048 & 2 & 64 & 128 & Fit & 89 & Qwen3-1.7B & 1,409,410,048 & 2 & 64 & 1024 & Fit \\
90 & Qwen3-1.7B & 1,409,410,048 & 2 & 128 & 8 & Fit & 91 & Qwen3-1.7B & 1,409,410,048 & 2 & 128 & 32 & Fit \\
92 & Qwen3-1.7B & 1,409,410,048 & 2 & 128 & 128 & Fit & 93 & Qwen3-1.7B & 1,409,410,048 & 2 & 128 & 1024 & Fit \\
94 & Qwen3-1.7B & 1,409,410,048 & 2 & 1024 & 8 & Fit & 95 & Qwen3-1.7B & 1,409,410,048 & 2 & 1024 & 32 & Fit \\
96 & Qwen3-1.7B & 1,409,410,048 & 2 & 1024 & 128 & Fit & 97 & Qwen3-1.7B & 1,409,410,048 & 2 & 1024 & 1024 & Fit \\
98 & Qwen3-4B & 3,633,511,936 & 8 & 128 & 128 & Fit & 99 & Qwen3-4B & 3,633,511,936 & 4 & 32 & 8 & Fit \\
100 & Qwen3-4B & 3,633,511,936 & 4 & 32 & 32 & Fit & 101 & Qwen3-4B & 3,633,511,936 & 4 & 32 & 128 & Fit \\
102 & Qwen3-4B & 3,633,511,936 & 4 & 32 & 1024 & Fit & 103 & Qwen3-4B & 3,633,511,936 & 4 & 64 & 8 & Fit \\
104 & Qwen3-4B & 3,633,511,936 & 4 & 64 & 32 & Fit & 105 & Qwen3-4B & 3,633,511,936 & 4 & 64 & 128 & Fit \\
106 & Qwen3-4B & 3,633,511,936 & 4 & 64 & 1024 & Fit & 107 & Qwen3-4B & 3,633,511,936 & 4 & 128 & 8 & Fit \\
108 & Qwen3-4B & 3,633,511,936 & 4 & 128 & 32 & Fit & 109 & Qwen3-4B & 3,633,511,936 & 4 & 128 & 128 & Fit \\
110 & Qwen3-4B & 3,633,511,936 & 4 & 128 & 1024 & Fit & 111 & Qwen3-4B & 3,633,511,936 & 4 & 1024 & 8 & Fit \\
112 & Qwen3-4B & 3,633,511,936 & 4 & 1024 & 32 & Fit & 113 & Qwen3-4B & 3,633,511,936 & 4 & 1024 & 128 & Fit \\
114 & Qwen3-4B & 3,633,511,936 & 4 & 1024 & 1024 & Fit & 115 & Qwen3-4B & 3,633,511,936 & 3 & 32 & 8 & Fit \\
116 & Qwen3-4B & 3,633,511,936 & 3 & 32 & 32 & Fit & 117 & Qwen3-4B & 3,633,511,936 & 3 & 32 & 128 & Fit \\
118 & Qwen3-4B & 3,633,511,936 & 3 & 32 & 1024 & Fit & 119 & Qwen3-4B & 3,633,511,936 & 3 & 64 & 8 & Fit \\
120 & Qwen3-4B & 3,633,511,936 & 3 & 64 & 32 & Fit & 121 & Qwen3-4B & 3,633,511,936 & 3 & 64 & 128 & Fit \\
122 & Qwen3-4B & 3,633,511,936 & 3 & 64 & 1024 & Fit & 123 & Qwen3-4B & 3,633,511,936 & 3 & 128 & 8 & Fit \\
124 & Qwen3-4B & 3,633,511,936 & 3 & 128 & 32 & Fit & 125 & Qwen3-4B & 3,633,511,936 & 3 & 128 & 128 & Fit \\
126 & Qwen3-4B & 3,633,511,936 & 3 & 128 & 1024 & Fit & 127 & Qwen3-4B & 3,633,511,936 & 3 & 1024 & 8 & Fit \\
128 & Qwen3-4B & 3,633,511,936 & 3 & 1024 & 32 & Fit & 129 & Qwen3-4B & 3,633,511,936 & 3 & 1024 & 128 & Fit \\
130 & Qwen3-4B & 3,633,511,936 & 3 & 1024 & 1024 & Fit & 131 & Qwen3-4B & 3,633,511,936 & 2 & 32 & 8 & Fit \\
132 & Qwen3-4B & 3,633,511,936 & 2 & 32 & 32 & Fit & 133 & Qwen3-4B & 3,633,511,936 & 2 & 32 & 128 & Fit \\
134 & Qwen3-4B & 3,633,511,936 & 2 & 32 & 1024 & Fit & 135 & Qwen3-4B & 3,633,511,936 & 2 & 64 & 8 & Fit \\
136 & Qwen3-4B & 3,633,511,936 & 2 & 64 & 32 & Fit & 137 & Qwen3-4B & 3,633,511,936 & 2 & 64 & 128 & Fit \\
138 & Qwen3-4B & 3,633,511,936 & 2 & 64 & 1024 & Fit & 139 & Qwen3-4B & 3,633,511,936 & 2 & 128 & 8 & Fit \\
140 & Qwen3-4B & 3,633,511,936 & 2 & 128 & 32 & Fit & 141 & Qwen3-4B & 3,633,511,936 & 2 & 128 & 128 & Fit \\
142 & Qwen3-4B & 3,633,511,936 & 2 & 128 & 1024 & Fit & 143 & Qwen3-4B & 3,633,511,936 & 2 & 1024 & 8 & Fit \\
144 & Qwen3-4B & 3,633,511,936 & 2 & 1024 & 32 & Fit & 145 & Qwen3-4B & 3,633,511,936 & 2 & 1024 & 128 & Fit \\
146 & Qwen3-4B & 3,633,511,936 & 2 & 1024 & 1024 & Fit & 147 & Qwen3-8B & 6,946,075,648 & 8 & 128 & 128 & Fit \\
148 & Qwen3-8B & 6,946,075,648 & 4 & 32 & 8 & Fit & 149 & Qwen3-8B & 6,946,075,648 & 4 & 32 & 32 & Fit \\
150 & Qwen3-8B & 6,946,075,648 & 4 & 32 & 128 & Fit & 151 & Qwen3-8B & 6,946,075,648 & 4 & 32 & 1024 & Fit \\
152 & Qwen3-8B & 6,946,075,648 & 4 & 64 & 8 & Fit & 153 & Qwen3-8B & 6,946,075,648 & 4 & 64 & 32 & Fit \\
154 & Qwen3-8B & 6,946,075,648 & 4 & 64 & 128 & Fit & 155 & Qwen3-8B & 6,946,075,648 & 4 & 64 & 1024 & Fit \\
156 & Qwen3-8B & 6,946,075,648 & 4 & 128 & 8 & Fit & 157 & Qwen3-8B & 6,946,075,648 & 4 & 128 & 32 & Fit \\
158 & Qwen3-8B & 6,946,075,648 & 4 & 128 & 128 & Fit & 159 & Qwen3-8B & 6,946,075,648 & 4 & 128 & 1024 & Fit \\
160 & Qwen3-8B & 6,946,075,648 & 4 & 1024 & 8 & Fit & 161 & Qwen3-8B & 6,946,075,648 & 4 & 1024 & 32 & Fit \\
162 & Qwen3-8B & 6,946,075,648 & 4 & 1024 & 128 & Fit & 163 & Qwen3-8B & 6,946,075,648 & 4 & 1024 & 1024 & Fit \\
164 & Qwen3-8B & 6,946,075,648 & 3 & 32 & 8 & Fit & 165 & Qwen3-8B & 6,946,075,648 & 3 & 32 & 32 & Fit \\
166 & Qwen3-8B & 6,946,075,648 & 3 & 32 & 128 & Fit & 167 & Qwen3-8B & 6,946,075,648 & 3 & 32 & 1024 & Fit \\
168 & Qwen3-8B & 6,946,075,648 & 3 & 64 & 8 & Fit & 169 & Qwen3-8B & 6,946,075,648 & 3 & 64 & 32 & Fit \\
170 & Qwen3-8B & 6,946,075,648 & 3 & 64 & 128 & Fit & 171 & Qwen3-8B & 6,946,075,648 & 3 & 64 & 1024 & Fit \\
172 & Qwen3-8B & 6,946,075,648 & 3 & 128 & 8 & Fit & 173 & Qwen3-8B & 6,946,075,648 & 3 & 128 & 32 & Fit \\
174 & Qwen3-8B & 6,946,075,648 & 3 & 128 & 128 & Fit & 175 & Qwen3-8B & 6,946,075,648 & 3 & 128 & 1024 & Fit \\
176 & Qwen3-8B & 6,946,075,648 & 3 & 1024 & 8 & Fit & 177 & Qwen3-8B & 6,946,075,648 & 3 & 1024 & 32 & Fit \\
178 & Qwen3-8B & 6,946,075,648 & 3 & 1024 & 128 & Fit & 179 & Qwen3-8B & 6,946,075,648 & 3 & 1024 & 1024 & Fit \\
180 & Qwen3-8B & 6,946,075,648 & 2 & 32 & 8 & Fit & 181 & Qwen3-8B & 6,946,075,648 & 2 & 32 & 32 & Fit \\
182 & Qwen3-8B & 6,946,075,648 & 2 & 32 & 128 & Fit & 183 & Qwen3-8B & 6,946,075,648 & 2 & 32 & 1024 & Fit \\
184 & Qwen3-8B & 6,946,075,648 & 2 & 64 & 8 & Fit & 185 & Qwen3-8B & 6,946,075,648 & 2 & 64 & 32 & Fit \\
186 & Qwen3-8B & 6,946,075,648 & 2 & 64 & 128 & Fit & 187 & Qwen3-8B & 6,946,075,648 & 2 & 64 & 1024 & Fit \\
188 & Qwen3-8B & 6,946,075,648 & 2 & 128 & 8 & Fit & 189 & Qwen3-8B & 6,946,075,648 & 2 & 128 & 32 & Fit \\
190 & Qwen3-8B & 6,946,075,648 & 2 & 128 & 128 & Fit & 191 & Qwen3-8B & 6,946,075,648 & 2 & 128 & 1024 & Fit \\
192 & Qwen3-8B & 6,946,075,648 & 2 & 1024 & 8 & Fit & 193 & Qwen3-8B & 6,946,075,648 & 2 & 1024 & 32 & Fit \\
194 & Qwen3-8B & 6,946,075,648 & 2 & 1024 & 128 & Fit & 195 & Qwen3-8B & 6,946,075,648 & 2 & 1024 & 1024 & Fit \\
196 & Qwen3-14B & 13,212,482,560 & 8 & 128 & 128 & Fit & 197 & Qwen3-14B & 13,212,482,560 & 4 & 32 & 8 & Fit \\
198 & Qwen3-14B & 13,212,482,560 & 4 & 32 & 32 & Fit & 199 & Qwen3-14B & 13,212,482,560 & 4 & 32 & 128 & Fit \\
200 & Qwen3-14B & 13,212,482,560 & 4 & 32 & 1024 & Fit & 201 & Qwen3-14B & 13,212,482,560 & 4 & 64 & 8 & Fit \\
202 & Qwen3-14B & 13,212,482,560 & 4 & 64 & 32 & Fit & 203 & Qwen3-14B & 13,212,482,560 & 4 & 64 & 128 & Fit \\
204 & Qwen3-14B & 13,212,482,560 & 4 & 64 & 1024 & Fit & 205 & Qwen3-14B & 13,212,482,560 & 4 & 128 & 8 & Fit \\
206 & Qwen3-14B & 13,212,482,560 & 4 & 128 & 32 & Fit & 207 & Qwen3-14B & 13,212,482,560 & 4 & 128 & 128 & Fit \\
208 & Qwen3-14B & 13,212,482,560 & 4 & 128 & 1024 & Fit & 209 & Qwen3-14B & 13,212,482,560 & 4 & 1024 & 8 & Fit \\
210 & Qwen3-14B & 13,212,482,560 & 4 & 1024 & 32 & Fit & 211 & Qwen3-14B & 13,212,482,560 & 4 & 1024 & 128 & Fit \\
212 & Qwen3-14B & 13,212,482,560 & 4 & 1024 & 1024 & Fit & 213 & Qwen3-14B & 13,212,482,560 & 3 & 32 & 8 & Fit \\
214 & Qwen3-14B & 13,212,482,560 & 3 & 32 & 32 & Fit & 215 & Qwen3-14B & 13,212,482,560 & 3 & 32 & 128 & Fit \\
216 & Qwen3-14B & 13,212,482,560 & 3 & 32 & 1024 & Fit & 217 & Qwen3-14B & 13,212,482,560 & 3 & 64 & 8 & Fit \\
218 & Qwen3-14B & 13,212,482,560 & 3 & 64 & 32 & Fit & 219 & Qwen3-14B & 13,212,482,560 & 3 & 64 & 128 & Fit \\
220 & Qwen3-14B & 13,212,482,560 & 3 & 64 & 1024 & Fit & 221 & Qwen3-14B & 13,212,482,560 & 3 & 128 & 8 & Fit \\
222 & Qwen3-14B & 13,212,482,560 & 3 & 128 & 32 & Fit & 223 & Qwen3-14B & 13,212,482,560 & 3 & 128 & 128 & Fit \\
224 & Qwen3-14B & 13,212,482,560 & 3 & 128 & 1024 & Fit & 225 & Qwen3-14B & 13,212,482,560 & 3 & 1024 & 8 & Fit \\
226 & Qwen3-14B & 13,212,482,560 & 3 & 1024 & 32 & Fit & 227 & Qwen3-14B & 13,212,482,560 & 3 & 1024 & 128 & Fit \\
228 & Qwen3-14B & 13,212,482,560 & 3 & 1024 & 1024 & Fit & 229 & Qwen3-14B & 13,212,482,560 & 2 & 32 & 8 & Fit \\
230 & Qwen3-14B & 13,212,482,560 & 2 & 32 & 32 & Fit & 231 & Qwen3-14B & 13,212,482,560 & 2 & 32 & 128 & Fit \\
232 & Qwen3-14B & 13,212,482,560 & 2 & 32 & 1024 & Fit & 233 & Qwen3-14B & 13,212,482,560 & 2 & 64 & 8 & Fit \\
234 & Qwen3-14B & 13,212,482,560 & 2 & 64 & 32 & Fit & 235 & Qwen3-14B & 13,212,482,560 & 2 & 64 & 128 & Fit \\
236 & Qwen3-14B & 13,212,482,560 & 2 & 64 & 1024 & Fit & 237 & Qwen3-14B & 13,212,482,560 & 2 & 128 & 8 & Fit \\
238 & Qwen3-14B & 13,212,482,560 & 2 & 128 & 32 & Fit & 239 & Qwen3-14B & 13,212,482,560 & 2 & 128 & 128 & Fit \\
240 & Qwen3-14B & 13,212,482,560 & 2 & 128 & 1024 & Fit & 241 & Qwen3-14B & 13,212,482,560 & 2 & 1024 & 8 & Fit \\
242 & Qwen3-14B & 13,212,482,560 & 2 & 1024 & 32 & Fit & 243 & Qwen3-14B & 13,212,482,560 & 2 & 1024 & 128 & Fit \\
244 & Qwen3-14B & 13,212,482,560 & 2 & 1024 & 1024 & Fit & 245 & Qwen3-32B & 31,206,298,624 & 8 & 128 & 128 & Val \\
246 & Qwen3-32B & 31,206,298,624 & 4 & 32 & 128 & Val & 247 & Qwen3-32B & 31,206,298,624 & 4 & 128 & 8 & Val \\
248 & Qwen3-32B & 31,206,298,624 & 4 & 128 & 128 & Val & 249 & Qwen3-32B & 31,206,298,624 & 4 & 1024 & 128 & Val \\
250 & Qwen3-32B & 31,206,298,624 & 3 & 128 & 128 & Val & 251 & Llama-3.2-1B & 973,146,112 & 4 & 32 & 128 & Gen \\
252 & Llama-3.2-1B & 973,146,112 & 4 & 64 & 128 & Gen & 253 & Llama-3.2-1B & 973,146,112 & 4 & 128 & 8 & Gen \\
254 & Llama-3.2-1B & 973,146,112 & 4 & 128 & 32 & Gen & 255 & Llama-3.2-1B & 973,146,112 & 4 & 128 & 128 & Gen \\
256 & Llama-3.2-1B & 973,146,112 & 4 & 128 & 1024 & Gen & 257 & Llama-3.2-1B & 973,146,112 & 4 & 1024 & 128 & Gen \\
258 & Llama-3.2-1B & 973,146,112 & 3 & 32 & 128 & Gen & 259 & Llama-3.2-1B & 973,146,112 & 3 & 64 & 128 & Gen \\
260 & Llama-3.2-1B & 973,146,112 & 3 & 128 & 8 & Gen & 261 & Llama-3.2-1B & 973,146,112 & 3 & 128 & 32 & Gen \\
262 & Llama-3.2-1B & 973,146,112 & 3 & 128 & 128 & Gen & 263 & Llama-3.2-1B & 973,146,112 & 3 & 128 & 1024 & Gen \\
264 & Llama-3.2-1B & 973,146,112 & 3 & 1024 & 128 & Gen & 265 & Llama-3.2-3B & 2,818,747,392 & 4 & 32 & 128 & Gen \\
266 & Llama-3.2-3B & 2,818,747,392 & 4 & 64 & 128 & Gen & 267 & Llama-3.2-3B & 2,818,747,392 & 4 & 128 & 8 & Gen \\
268 & Llama-3.2-3B & 2,818,747,392 & 4 & 128 & 32 & Gen & 269 & Llama-3.2-3B & 2,818,747,392 & 4 & 128 & 128 & Gen \\
270 & Llama-3.2-3B & 2,818,747,392 & 4 & 128 & 1024 & Gen & 271 & Llama-3.2-3B & 2,818,747,392 & 4 & 1024 & 128 & Gen \\
272 & Llama-3.2-3B & 2,818,747,392 & 3 & 32 & 128 & Gen & 273 & Llama-3.2-3B & 2,818,747,392 & 3 & 64 & 128 & Gen \\
274 & Llama-3.2-3B & 2,818,747,392 & 3 & 128 & 8 & Gen & 275 & Llama-3.2-3B & 2,818,747,392 & 3 & 128 & 32 & Gen \\
276 & Llama-3.2-3B & 2,818,747,392 & 3 & 128 & 128 & Gen & 277 & Llama-3.2-3B & 2,818,747,392 & 3 & 128 & 1024 & Gen \\
278 & Llama-3.2-3B & 2,818,747,392 & 3 & 1024 & 128 & Gen & 279 & Llama-3.1-8B & 6,979,588,096 & 4 & 32 & 128 & Gen \\
280 & Llama-3.1-8B & 6,979,588,096 & 4 & 64 & 128 & Gen & 281 & Llama-3.1-8B & 6,979,588,096 & 4 & 128 & 8 & Gen \\
282 & Llama-3.1-8B & 6,979,588,096 & 4 & 128 & 32 & Gen & 283 & Llama-3.1-8B & 6,979,588,096 & 4 & 128 & 128 & Gen \\
284 & Llama-3.1-8B & 6,979,588,096 & 4 & 128 & 1024 & Gen & 285 & Llama-3.1-8B & 6,979,588,096 & 4 & 1024 & 128 & Gen \\
286 & Llama-3.1-8B & 6,979,588,096 & 3 & 32 & 128 & Gen & 287 & Llama-3.1-8B & 6,979,588,096 & 3 & 64 & 128 & Gen \\
288 & Llama-3.1-8B & 6,979,588,096 & 3 & 128 & 8 & Gen & 289 & Llama-3.1-8B & 6,979,588,096 & 3 & 128 & 32 & Gen \\
290 & Llama-3.1-8B & 6,979,588,096 & 3 & 128 & 128 & Gen & 291 & Llama-3.1-8B & 6,979,588,096 & 3 & 128 & 1024 & Gen \\
292 & Llama-3.1-8B & 6,979,588,096 & 3 & 1024 & 128 & Gen & & & & & & & \\
\end{longtable}
\end{small}
\twocolumn 

\end{document}